\documentclass{article}

\usepackage{microtype}
\usepackage{graphicx}
\usepackage{subfigure}
\usepackage{booktabs} 

\usepackage{hyperref}

\usepackage[accepted]{icml2025}

\usepackage{amsmath}
\usepackage{amssymb}
\usepackage{mathtools}
\usepackage{amsthm}

\usepackage{algorithm}
\usepackage{algorithmic}

\usepackage[capitalize,noabbrev]{cleveref}

\usepackage{float}

\theoremstyle{plain}

\theoremstyle{definition}

\theoremstyle{remark}

\usepackage[textsize=tiny]{todonotes}

\icmltitlerunning{FedSCAM: Scam-resistant SAM for Robust Federated Optimization in Heterogeneous Environments}

\begin{document}

\twocolumn[
\icmltitle{FedSCAM: Scam-resistant SAM for Robust Federated Optimization \\
in Heterogeneous Environments}

\icmlsetsymbol{equal}{*}

\begin{icmlauthorlist}
\icmlauthor{Sameer Rahil}{lums}
\icmlauthor{Zain Abdullah Ahmad}{lums}
\icmlauthor{Talha Asif}{lums}
\end{icmlauthorlist}

\icmlaffiliation{lums}{Department of Computer Science, Lahore University of Management Sciences, Lahore, Pakistan}

\icmlcorrespondingauthor{Sameer Rahil}{27100162@lums.edu.pk}
\icmlcorrespondingauthor{Zain Abdullah Ahmad}{27100215@lums.edu.pk}
\icmlcorrespondingauthor{Talha Asif}{24280053@lums.edu.pk}

\icmlkeywords{Federated Learning, FedAvg, Dirichlet partition, label skew, heterogeneity, FedProx, SCAFFOLD, FedGH, FedSAM}

\vskip 0.3in
]

\printAffiliationsAndNotice{\icmlEqualContribution}

\begin{abstract}
Federated Learning (FL) enables collaborative model training across decentralized edge devices while preserving data privacy. However, statistical heterogeneity among clients, often manifested as non-IID label distributions, poses significant challenges to convergence and generalization. While Sharpness-Aware Minimization (SAM) has been introduced to FL to seek flatter, more robust minima, existing approaches typically apply a uniform perturbation radius across all clients, ignoring client-specific heterogeneity. In this work, we propose \textbf{FedSCAM} (Federated Sharpness-Aware Minimization with Clustered Aggregation and Modulation), a novel algorithm that dynamically adjusts the SAM perturbation radius and aggregation weights based on client-specific heterogeneity scores. By calculating a heterogeneity metric for each client and modulating the perturbation radius inversely to this score, FedSCAM prevents clients with high variance from destabilizing the global model. Furthermore, we introduce a heterogeneity-aware weighted aggregation mechanism that prioritizes updates from clients that align with the global optimization direction. Extensive experiments on CIFAR-10 and Fashion-MNIST under various degrees of Dirichlet-based label skew demonstrate that FedSCAM achieves competitive performance among state-of-the-art baselines, including FedSAM, FedLESAM, etc. in terms of convergence speed and final test accuracy. Code and artifacts available at: \href{https://github.com/SameerRahil/FedSCAM}{Github Repository}.
\end{abstract}

\section{Introduction}
\label{introduction}
The proliferation of smart devices and the increasing concern for data privacy have catalyzed the adoption of Federated Learning (FL) as a paradigm for distributed machine learning. In the canonical FL setting, a central server orchestrates the training process by distributing a global model to a subset of clients, who then perform local training on their private data and return model updates for aggregation. The server aggregates these updates, typically using Federated Averaging (FedAvg) \cite{mcmahan2017fedavg}, to update the global model. This process repeats until convergence.

While FL offers a promising solution to privacy constraints, it introduces new challenges not present in centralized training, primarily stemming from the statistical heterogeneity of client data. In real-world scenarios, client data is rarely independent and identically distributed (IID). Instead, clients often possess highly skewed class distributions, referred to as label skew. This statistical heterogeneity causes the local objective functions of clients to diverge significantly from the global objective function. Consequently, local updates computed by clients may point in conflicting directions, leading to "client drift" where the aggregated global model struggles to converge or settles into a suboptimal minimum. Standard optimization techniques like Stochastic Gradient Descent (SGD) are insufficient to handle this drift effectively, necessitating the development of algorithms specifically designed for the heterogeneous FL landscape.

Recent advancements in centralized optimization, particularly Sharpness-Aware Minimization (SAM) \cite{foret2021sharpness}, have shown that models converging to "flat" minima - regions where the loss function is relatively invariant to small perturbations in parameters - generalize better than those converging to "sharp" minima. This insight has been adapted to FL in the form of FedSAM \cite{qu2022fedsam}, which encourages clients to find flat local minima, theoretically improving the robustness of the aggregated global model. However, standard FedSAM and its variants apply a static perturbation radius $\rho$ across all clients. We argue that this "one-size-fits-all" approach is suboptimal in heterogeneous settings. Clients with high data heterogeneity or "noisy" local landscapes should arguably be subjected to smaller perturbations to prevent divergence, while clients with cleaner, more representative data can afford larger perturbations to explore the landscape more aggressively.

To address this limitation, we propose \textbf{FedSCAM}, a heterogeneity-aware framework that modulates both the local optimization process and the global aggregation strategy. FedSCAM introduces a dynamic mechanism to calculate a client-specific heterogeneity score based on the gradient norms of local batches. This score is then used to adaptively scale the SAM perturbation radius for each client. Clients exhibiting high heterogeneity are constrained with a smaller radius, while stable clients utilize a larger radius. Furthermore, FedSCAM eschews the traditional sample-size-weighted averaging of FedAvg in favor of a weighted aggregation scheme that penalizes clients with high heterogeneity scores and rewards clients whose updates align with the trajectory of the global model.

Our contributions are as follows:
\begin{itemize}
    \item We propose an adaptive radius mechanism for SAM in Federated Learning, where the perturbation magnitude is inversely proportional to a dynamically computed client heterogeneity score.
    \item We design a novel aggregation strategy that weights client contributions based on their heterogeneity score and their alignment with the global model update, reducing the impact of outliers.
    \item We conduct an empirical evaluation on CIFAR-10 and Fashion-MNIST under Dirichlet label-skew partitions ($\alpha \in \{0.1, 0.5, 1.0\}$ where smaller $\alpha$ is more non-IID), using ResNet-18 and a lightweight CNN where applicable.
    \item We demonstrate that FedSCAM is competitive against strong baselines we implemented and ran under identical splits and hyperparameters, including FedAvg-style aggregation baselines (FedAvgM, q-FedAvg, FedLW, FedNoLoWe) and SAM-family baselines (FedSAM, FedLESAM, FedWMSAM), as well as the hybrid diagnostic baseline FedLWSAM.

\end{itemize}

The remainder of this paper is organized as follows: Section \ref{sec:related} reviews related work and the baselines used in this study. Section \ref{sec:algorithm} details the proposed FedSCAM algorithm. Section \ref{sec:methodology} describes the experimental setup and implementation details. Section \ref{sec:results} presents the results and sensitivity analysis. Finally, Section \ref{sec:discussion} and \ref{sec:conclusion} discuss the implications of our findings and conclude the paper.


\section{Related Work}
\label{sec:related}

Federated optimization methods under label-skew heterogeneity typically improve performance by (i) stabilizing aggregation against conflicting updates, or (ii) improving local objectives to generalize better. FedSCAM sits at their intersection: we modulate a flatness-seeking local optimizer (SAM) on a per-client basis and simultaneously reweight aggregation using client-specific reliability and alignment signals.

\subsection{Aggregation and Client Reweighting under Heterogeneity}
\textbf{FedAvg}~\cite{mcmahan2017fedavg} is the canonical FL baseline: each selected client performs local SGD for a few epochs and the server aggregates the resulting updates using weights proportional to the client sample counts. This simple averaging works well under near-IID data, but under label skew or feature shift, different clients may produce gradients pointing in conflicting directions. As a result, the global model can oscillate across rounds and may converge to a solution that overfits dominant client distributions rather than improving broad generalization.

A practical stabilization is \textbf{FedAvgM}~\cite{hsu2019measuring}, which adds \emph{server-side momentum} to smooth the sequence of aggregated updates. Intuitively, momentum acts as a low-pass filter: it dampens high-variance update noise induced by heterogeneous client objectives and reduces round-to-round zig-zagging. This is especially helpful when local training is aggressive (more local steps) or when participation is partial, both of which can increase update variance.

\textbf{q-FedAvg}~\cite{li2020qfedavg} addresses heterogeneity from a \emph{fairness} perspective by reweighting clients according to their loss. Instead of treating the goal as minimizing the average client objective, it emphasizes clients that currently incur larger losses, thereby reducing performance disparity across clients. In practice, this reweighting can shift training toward under-served or harder distributions, which may improve worst-client accuracy but can also alter the global optimization trajectory relative to FedAvg.

\textbf{FedProx}~\cite{li2020fedprox} tackles ``client drift'' by modifying the local objective with a proximal regularizer that penalizes deviation from the current global model. Concretely, each client optimizes its local loss plus a term of the form $\frac{\mu}{2}\|w-w^{(t)}\|_2^2$, where $w^{(t)}$ is the round-$t$ global model. This anchors local updates when data are highly non-IID or when clients have variable compute (unequal local steps), improving stability and making local solutions less likely to over-specialize to a client’s distribution.

Loss-based reweighting is another lightweight approach: \textbf{FedLW}~\cite{yao2025fedlw} assigns higher aggregation weight to clients using a training-loss-derived reliability signal, aiming to downweight noisy or poorly optimized client updates. We additionally report \textbf{FedNoLoWe}~\cite{le2025fednolowe}, a normalized loss-based weighting strategy that stabilizes the weighting signal by normalization, reducing sensitivity to scale differences across clients or rounds. Both methods can be viewed as ``plug-in'' alternatives to sample-count weighting that attempt to make aggregation more robust under heterogeneity without changing the client optimizer.

Finally, we report a hybrid diagnostic baseline \textbf{FedLWSAM}, which combines FedLW-style loss weighting with SAM-based local optimization. This composition baseline isolates whether gains stem primarily from (i) improved aggregation weights, or (ii) sharpness-aware local training. Concretely, clients perform SAM updates locally, and the server aggregates client deltas using the same loss-based weighting rule as FedLW.

\subsection{Sharpness-Aware Minimization in Federated Learning}
\textbf{SAM}~\cite{foret2021sharpness} improves generalization by explicitly seeking \emph{flat} minima, optimizing a robust objective that minimizes the worst-case loss in a small neighborhood around the current weights:
\begin{equation}
\min_{w}\;\max_{\|\epsilon\|_2 \le \rho}\; L(w+\epsilon).
\end{equation}
Operationally, SAM performs a two-step update: it first perturbs weights in the direction of the gradient (scaled to have norm $\rho$), then takes a descent step using the gradient evaluated at the perturbed weights. By discouraging solutions that are highly sensitive to small parameter perturbations, SAM often yields models that generalize better than standard ERM training.

\textbf{FedSAM}~\cite{qu2022fedsam} brings SAM into FL by running SAM-based local updates at clients and then aggregating model deltas at the server, mirroring FedAvg’s communication pattern. A key limitation under heterogeneous FL is that FedSAM typically uses a \emph{single global} perturbation radius $\rho$ for all clients, despite clients differing substantially in gradient scale, noise, and local curvature. In such settings, a uniform $\rho$ can under-regularize unstable clients or over-regularize stable ones, leading to suboptimal robustness--accuracy trade-offs across the federation.

Efficiency-oriented SAM-in-FL methods include \textbf{FedLESAM}~\cite{jiang2024fedlesam}, which reduces SAM overhead by reusing or estimating a consistent perturbation direction so that the extra backward/forward cost is mitigated, and \textbf{FedWMSAM}~\cite{li2025fedwmsam}, which injects momentum into the perturbation mechanism to smooth the perturbation dynamics across steps. These methods target the practical bottleneck of SAM in FL - extra computation per local step - while attempting to retain the generalization benefits of sharpness-aware training.

\paragraph{Positioning of FedSCAM.}
FedSCAM differs from the above by \emph{client-wise modulation}: we (i) estimate a lightweight client heterogeneity score from early-batch gradient norms, (ii) adapt each client’s SAM radius accordingly (rather than using a single global $\rho$), and (iii) perform heterogeneity- and alignment-aware aggregation, optionally with a clustered conflict-dampening step using low-dimensional update summaries. In our implementation, the client-specific pilot radius is set as $\rho_{\mathrm{pilot},i} \leftarrow \frac{0.5\,\rho_{\max}}{1+\alpha_\rho h_i}$, which provides a conservative per-client initialization before full client-wise modulation.

\section{The FedSCAM Algorithm}
\label{sec:algorithm}

\paragraph{Notation.}
Let $w_t$ denote the global model at round $t$, and let $\mathcal{S}_t$ be the set of selected clients. Each client $i$ has local dataset size $N_i$ and local data $\mathcal{D}_i$. Clients return model updates $\Delta_i = w_i - w_t$.

\paragraph{Implementation overview.}
FedSCAM has three core components per round:
(1) \emph{Heterogeneity estimation} using (a small number of) gradient-norm measurements;
(2) \emph{Adaptive SAM} where each client uses its own $\rho_i$;
(3) \emph{Heterogeneity- and alignment-aware aggregation}, optionally preceded by a light clustered conflict-dampening step.

\paragraph{Key design choices: Client-wise radius modulation.} In heterogeneous regimes, raw gradient scale and update variance differ across clients. A fixed $\rho$ can over-perturb unstable clients, amplifying drift. FedSCAM shrinks $\rho_i$ when $h_i^{\mathrm{adj}}$ is large, acting as a per-client ``trust throttle.''

\textbf{Alignment-aware adjustment.} Not all ``large gradients'' are harmful: if a client’s pilot direction aligns with the global direction, it likely contributes useful signal even under skew. By reducing $h_i^{\mathrm{adj}}$ when $c_i>0$, FedSCAM avoids over-penalizing helpful clients.

\textbf{Clustered conflict dampening.} When updates naturally group (e.g., due to similar label subsets), contradictory pairs can arise and cancel progress. The clustering step operates on low-dimensional summaries and suppresses severe within-cluster conflicts cheaply.

\begin{algorithm}[t]
\caption{FedSCAM (implementation-faithful pseudocode)}
\label{alg:fedscam}
\begin{algorithmic}
\small
\STATE {\bfseries Input:} Rounds $T$, local epochs $E$, max radius $\rho_{\max}$, radius scale $\alpha_\rho$, heterogeneity penalty $\gamma$, alignment boost $\beta$, heterogeneity-alignment coupling $\kappa$, conflict downweight $\lambda\in(0,1]$, clusters $K$, summary dim $d$.
\STATE Initialize global model $w_0$. Initialize global direction summary $u_0 \leftarrow \mathbf{0}$.

\FOR{$t=0$ {\bfseries to} $T-1$}
    \STATE Server selects clients $\mathcal{S}_t$ and broadcasts $w_t$ (and optional metadata).
    \FOR{{\bfseries each} client $i \in \mathcal{S}_t$ {\bfseries in parallel}}
        \STATE \raggedright \textbf{(Heterogeneity)} Using first $B$ batches, compute $h_i \approx \frac{1}{B}\sum_{b=1}^{B}\|\nabla L(w_t;\mathcal{B}_{i,b})\|_2$
        \STATE \textbf{(Pilot direction)} Compute a low-cost pilot direction $v_i$ on one batch (e.g., projected gradient or one-step update), then summarize: $s_i \leftarrow \mathrm{Proj}_d(\mathrm{Normalize}(v_i))$.
        \STATE \textbf{(Alignment)} $c_i \leftarrow \cos(s_i, u_{t-1})$ (if $u_{t-1}$ is undefined at $t=0$, set $c_i=0$).
        \STATE \textbf{(Adjusted heterogeneity)} $h_i^{\mathrm{adj}} \leftarrow h_i\cdot \max(0,\,1-\kappa c_i)$.
        \STATE \textbf{(Adaptive radius)} $\rho_i \leftarrow \frac{\rho_{\max}}{1+\alpha_\rho h_i^{\mathrm{adj}}}$.
        \STATE \textbf{(Local training)} Run SAM with radius $\rho_i$ for $E$ epochs starting from $w_t$ to obtain $w_i$; set $\Delta_i \leftarrow w_i - w_t$.
        \STATE \textbf{(Summary)} $z_i \leftarrow \mathrm{Proj}_d(\mathrm{Normalize}(\Delta_i))$.
    \ENDFOR

    \STATE \textbf{(Optional clustered conflict dampening)} Cluster $\{z_i\}_{i\in\mathcal{S}_t}$ into $K$ clusters; within each cluster, if two updates have negative cosine similarity, downweight the smaller-norm update by factor $\lambda$.
    \STATE \textbf{(Aggregation weights)} For each client $i$, set
    \STATE \hspace{1.2em}$S_i \propto N_i \cdot \frac{1}{1+\gamma h_i^{\mathrm{adj}}}\cdot \max(0,1+\beta c_i).$
    \STATE Aggregate: $w_{t+1} \leftarrow w_t + \sum_{i\in\mathcal{S}_t}\frac{S_i}{\sum_{j\in\mathcal{S}_t}S_j}\,\Delta_i$.
    \STATE Update direction memory $u_t \leftarrow \mathrm{Proj}_d(\mathrm{Normalize}(w_{t+1}-w_t))$.
\ENDFOR
\end{algorithmic}
\end{algorithm}

\section{Methodology and Experiments}
\label{sec:methodology}

We evaluate FedSCAM in supervised image classification under controlled label-skew heterogeneity. Our goal is to understand when heterogeneity-aware SAM modulation and aggregation help, and what the compute/accuracy trade-offs look like against SAM-focused and aggregation-focused baselines.

\subsection{Experimental Setup}

\subsubsection{Datasets.}
We evaluate on CIFAR-10 and Fashion-MNIST (FMNIST), two widely used image classification benchmarks that have become standard testbeds for federated optimization under statistical heterogeneity. Using these datasets serves two purposes: (i) they provide complementary difficulty and visual structure (natural images vs.\ grayscale apparel), and (ii) they enable direct, apples-to-apples comparison with prior FL and FedSAM-family baselines, many of which report results on the same benchmarks. We simulate label-skew non-IID partitions using a Dirichlet distribution over class proportions, $Dir(\alpha)$, where smaller $\alpha$ produces more skewed (more heterogeneous) client label distributions. In our experiments, $\alpha=0.1$ corresponds to extreme heterogeneity, $\alpha=0.5$ to moderate heterogeneity, and $\alpha=1.0$ to mild heterogeneity (near-IID relative to $\alpha=0.1$). For sanity checks and visual diagnostics, we also include an effectively IID-like configuration (e.g., $\alpha=1$) to verify that the partitioning procedure behaves as expected. We enforce a small minimum per-client sample count to avoid degenerate clients (see Appendix).

\subsubsection{Models}
We use two architectures to cover both lightweight and deeper regimes.

\emph{ResNet-18.}
We use the standard ResNet-18 design (as commonly used for CIFAR/FMNIST variants): an initial stem (convolution + normalization + nonlinearity), followed by four residual stages with two basic residual blocks per stage, and a final global average pooling + linear classifier. Concretely, the network comprises:
(i) a first convolutional layer (\texttt{conv} $\rightarrow$ \texttt{BN} $\rightarrow$ \texttt{ReLU}),
(ii) residual Stage 1 with 2 basic blocks at 64 channels,
(iii) residual Stage 2 with 2 basic blocks at 128 channels (downsampling on the first block),
(iv) residual Stage 3 with 2 basic blocks at 256 channels (downsampling on the first block),
(v) residual Stage 4 with 2 basic blocks at 512 channels (downsampling on the first block),
and (vi) global average pooling followed by a fully-connected layer producing logits for 10 classes. Each basic block uses two $3\times 3$ convolutions with identity (or projection) skip connections.

\emph{SmallCNN.}
For fast iteration and controlled ablations, we use a compact convolutional network consisting of a small stack of convolutional blocks followed by a lightweight MLP head. Specifically, SmallCNN uses repeated \texttt{Conv2D} layers with normalization and ReLU activations, interleaved with spatial downsampling (max-pooling), and ends with a flattened feature vector passed through one or two fully-connected layers to produce 10-way logits. This architecture is intentionally shallow to reduce local training cost per round while still being expressive enough to expose optimizer/aggregation differences under label skew. (The exact layer widths and kernel sizes match our released code and configuration files.)

\subsubsection{Training configuration}
Unless otherwise stated, we use $M=10$ clients with full participation ($C=1.0$) to isolate optimization effects from partial participation. For the Fashion-MNIST study (the main SAM-family comparison in this paper), we run $T=10$ communication rounds with $E=5$ local epochs, batch size 256, and SGD learning rate $\eta=0.01$. Across experiments, we vary the number of rounds and sometimes the local epochs to control compute and to match the evaluation protocol of the corresponding figures/tables. Unless a figure caption states otherwise, the base configuration is:
clients $M=10$ with $C=1.0$, rounds $T \in \{10,15,30\}$, local epochs $E$ as specified per experiment (e.g., $E=5$ on FMNIST and smaller $E$ for deeper models when controlling runtime), batch size 64 (unless explicitly set to 256 for FMNIST), and SGD with learning rate $\eta=0.01$.

\subsubsection{Compute accounting}
SAM-based local training typically requires \emph{two} forward/backward passes per minibatch. FedSCAM adds small additional overhead from: (i) heterogeneity estimation over the first $B$ batches (e.g., $B=3$), and (ii) a pilot-direction computation used to estimate alignment signals. We therefore report wall-clock round time where relevant (see \Cref{fig:fmnist-fedscam-time}) to contextualize comparisons with efficiency-oriented baselines such as FedLESAM.

\subsubsection{Fedscam Hyperparameters}
FedSCAM introduces a small set of interpretable hyperparameters that control (a) how aggressively the SAM radius is modulated per client, and (b) how strongly heterogeneous or misaligned updates are downweighted at aggregation.
We use $\rho_{\max}=0.05$ as the maximum allowable perturbation radius, and set each client radius via $\rho_i=\rho_{\max}/(1+\alpha_\rho h_i^{\mathrm{adj}})$ (with alignment-adjusted heterogeneity $h_i^{\mathrm{adj}}$).
The parameter $\alpha_\rho$ controls the \emph{strength of radius modulation}: larger $\alpha_\rho$ shrinks $\rho_i$ more aggressively for high-heterogeneity clients, which can improve stability under extreme label skew (e.g., $\alpha=0.1$) but may over-regularize and slow progress when heterogeneity is mild. In more heterogeneous regimes, increasing $\alpha_\rho$ is typically safer; in near-IID settings, smaller $\alpha_\rho$ often suffices.

For aggregation, $\gamma$ controls the \emph{heterogeneity penalty} in the client weight (larger $\gamma$ downweights high-$h_i^{\mathrm{adj}}$ clients more strongly), while $\beta$ controls the \emph{alignment boost} (larger $\beta$ upweights clients whose pilot direction aligns with the global direction). In highly non-IID settings, increasing $\gamma$ can prevent noisy clients from dominating aggregation, and a moderate positive $\beta$ helps preserve useful signal from clients that are heterogeneous but still directionally consistent with global progress. The coupling parameter $\kappa$ governs how strongly alignment reduces the effective heterogeneity score (i.e., how much ``credit'' aligned clients receive even if their raw gradient norms are large). When client drift is severe, higher $\kappa$ can prevent over-penalizing helpful but high-gradient clients; when alignment signals are noisy, a smaller $\kappa$ is safer.

Finally, the optional clustered conflict-dampening mechanism uses $K$ clusters and a within-cluster downweight factor $\lambda\in(0,1]$. Increasing $K$ yields finer grouping (potentially better separation of update modes under label skew) at slightly higher overhead, while smaller $\lambda$ more aggressively suppresses conflicting updates inside clusters. This step is most useful under extreme heterogeneity where update directions naturally form modes; in milder settings it can be disabled with negligible impact. Low-dimensional summaries use random projection with $d\in\{256,512\}$ depending on model size: larger $d$ preserves more directional information but increases minor overhead.

\subsubsection{Reproducibility note}
All methods are evaluated under identical data partitions and training settings (same client splits, $T$, $E$, batch size, and learning rate) to ensure fair comparison. Experiments were repeated twice (and in some cases three times) with different random seeds; while we do not report mean$\pm$std in the current tables, the observed trends are consistent across these reruns. To support full reproducibility, we release code and configuration files so that results can be rerun directly and additional seeds can be added straightforwardly.

\subsection{Baselines}
We group baselines by what they isolate:

\subsubsection{Category A: SAM-family (flatness-focused).}
\begin{enumerate}
    \item \textbf{FedSAM} \cite{qu2022fedsam}: local SAM with fixed $\rho$ and FedAvg aggregation.
    \item \textbf{FedLESAM} \cite{jiang2024fedlesam}: consistent/global perturbation direction estimation, designed to reduce SAM overhead.
    \item \textbf{FedWMSAM} \cite{li2025fedwmsam}: FedLESAM-style perturbation with global momentum smoothing.
\end{enumerate}

\subsubsection{Category B: Aggregation-focused (SGD local training).}
\begin{enumerate}
    \item \textbf{FedAvg} \cite{mcmahan2017fedavg},
    \item \textbf{Uniform} averaging,
    \item \textbf{FedAvgM} \cite{hsu2019measuring},
    \item \textbf{q-FedAvg} \cite{li2020qfedavg},
    \item \textbf{FedLW} \cite{yao2025fedlw},
    \item \textbf{FedNoLoWe} \cite{le2025fednolowe} (where included).
\end{enumerate}

\subsubsection{Hybrid/diagnostic baselines (ours)}
We report \textbf{FedLWSAM} as a practical hybrid baseline: FedLW-style weighting with SAM local steps. We also report FedSCAM ablations: \textbf{FedSCAM (WA)} for aggregation-only, \textbf{FedSCAM (SAM)} for radius-only, and \textbf{FedSCAM (Full)} for the combined method.

\section{Results}
\label{sec:results}

\subsection{Sensitivity Analysis of FedSCAM}
To understand the contribution of individual components, we analyze the sensitivity of FedSCAM to its three core hyperparameters on CIFAR-10 ($\alpha=0.1$). The detailed training dynamics and final accuracy comparisons are provided in the Appendix.

\subsubsection{Effect of Maximum Perturbation Radius ($\rho_{\max}$)}
We varied $\rho_{\max} \in \{0.01, 0.05, 0.1\}$. As shown in Figure \ref{fig:rhomax-final} (Appendix), the method is robust to larger radii. While $\rho_{\max}=0.05$ offers stable convergence, $\rho_{\max}=0.1$ achieved the highest final accuracy, suggesting that FedSCAM's adaptive modulation mechanism effectively prevents the instability typically associated with large perturbation radii in standard SAM. Conversely, an overly conservative radius ($\rho_{\max}=0.01$) limits the generalization benefits.

\subsubsection{Effect of Heterogeneity Penalty ($\gamma$)}
We evaluated the impact of down-weighting high-loss clients by varying $\gamma \in \{0, 1.0, 5.0\}$. Figure \ref{fig:gamma-final} (Appendix) demonstrates that introducing the heterogeneity penalty ($\gamma > 0$) consistently improves performance over the baseline ($\gamma=0$). The configuration $\gamma=1.0$ yielded the most stable gain, validating our hypothesis that reducing the aggregation weight of clients with extreme effective heterogeneity mitigates drift.

\subsubsection{Effect of Alignment Boost ($\beta$)}
We tested the alignment-based weighting boost with $\beta \in \{0, 0.8, 2.0\}$. Interestingly, in this specific low-heterogeneity regime ($\alpha=0.1$), aggressive up-weighting of aligned clients ($\beta > 0$) did not yield performance gains compared to the baseline $\beta=0$ (see Figure \ref{fig:beta-final}). This suggests that while detecting conflict is crucial (via $\gamma$), "rewarding" alignment may be redundant or potentially destabilizing when the data partition is not extremely skewed.

\begin{figure}
\centering
\includegraphics[width=\linewidth]{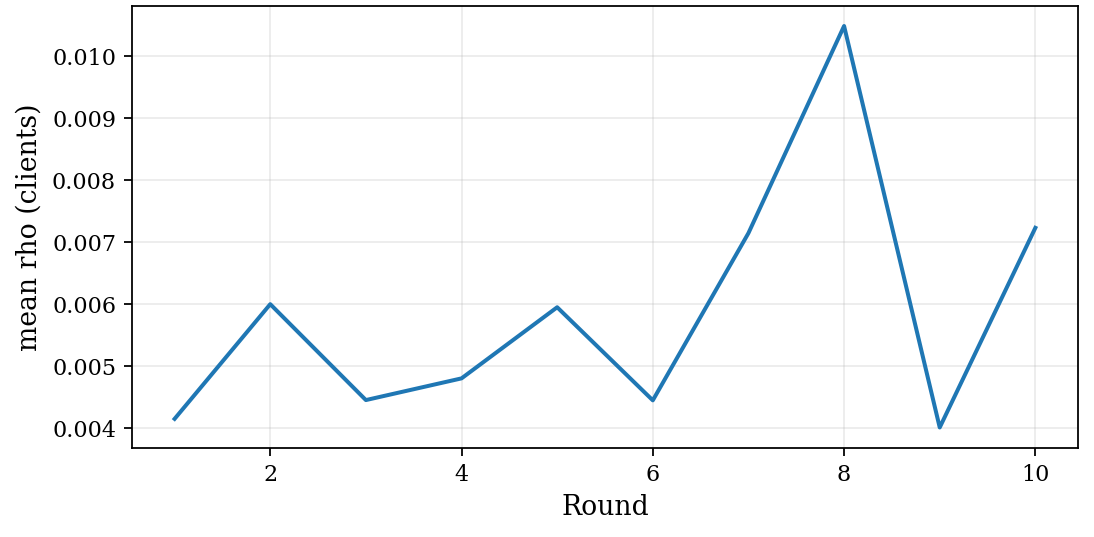}
  \vspace{-20pt}
\caption{Mean $\rho$ vs.\ communication rounds for FedSCAM on Fashion-MNIST, with $\rho _{max}$ set to 0.05}
\label{fig:fmnist-fedscam-rho}
\end{figure}

\subsection{Comparison with SAM Baselines}

\begin{figure}[t]
\centering
\includegraphics[width=\linewidth]{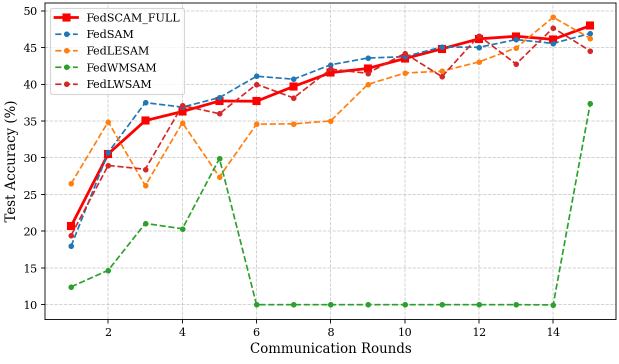}
  \vspace{-20pt}
\caption{SAM-family methods on CIFAR-10 under extreme heterogeneity ($\alpha=0.1$).}
\label{fig:cifar10-sam-alpha01}
\end{figure}

For the most heterogeneous setting ($\alpha = 0.1$), SAM-family baselines can behave quite differently depending on how perturbations are computed and how stable the aggregation is. On Fashion-MNIST with $\alpha=0.1$, \Cref{tab:fmnist-final-acc} reports final accuracies after 10 rounds. FedSCAM achieves \textbf{65.41\%}, very close to FedLESAM (\textbf{65.76\%}), and substantially above momentum-based FedWMSAM (\textbf{45.03\%}) and other loss-weighting hybrids reported in this setting.

\begin{figure}
\centering
\includegraphics[width=\linewidth]{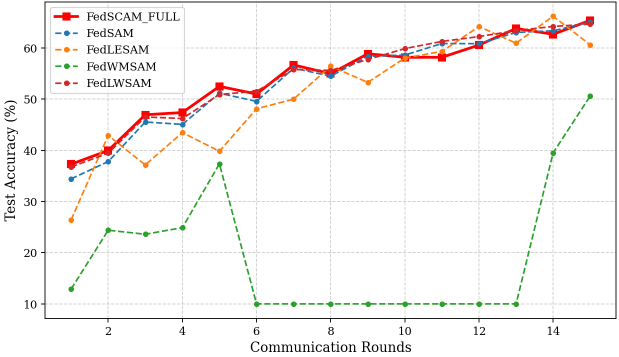}
  \vspace{-20pt}
\caption{SAM variants on CIFAR-10 ($\alpha=0.5$).}
\label{fig:cifar10-sam-alpha05}
\end{figure}

\begin{figure}
\centering
\includegraphics[width=\linewidth]{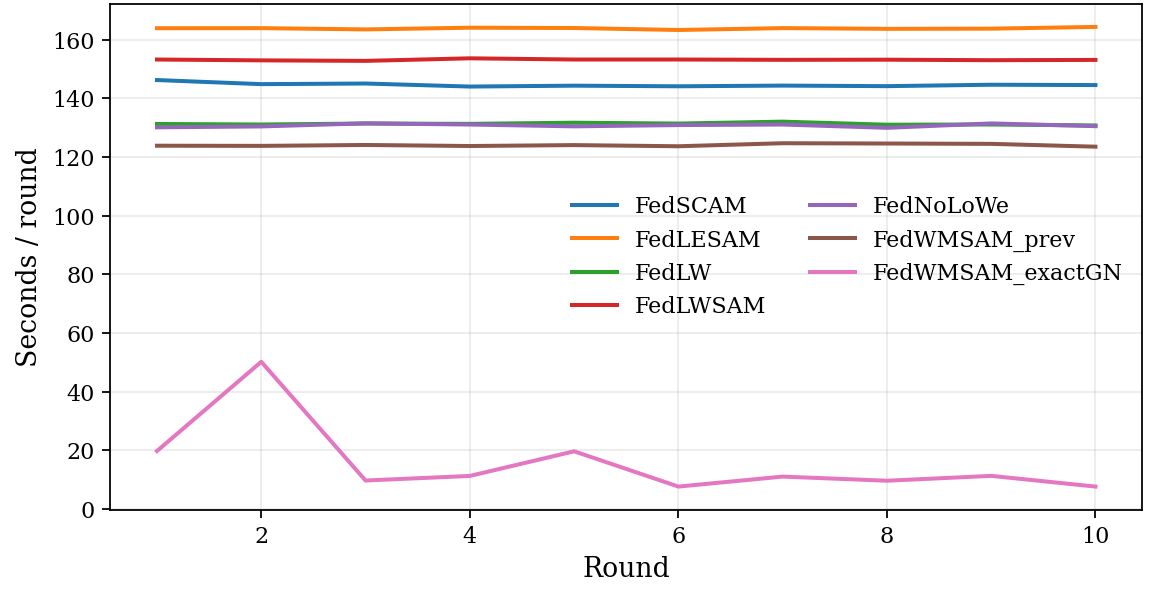}
  \vspace{-20pt}
\caption{Wall-clock time per communication round on Fashion-MNIST. FedSAM is faster than FedLESAM.}
\label{fig:fmnist-fedscam-time}
\end{figure}

A key practical difference is compute: in our runs, FedLESAM required roughly \textbf{$\sim$20 seconds more per round} than FedSCAM on the same hardware/configuration (see \Cref{fig:fmnist-fedscam-time} for the round-time comparison plot). This places FedSCAM in a favorable accuracy--efficiency regime: it matches FedLESAM closely while maintaining lower per-round compute.

We also observe that FedSCAM can provide consistent gains over FedSAM in moderate heterogeneity. For example, in a CIFAR-10 run at $\alpha=0.5$ (SmallCNN, 30 rounds), FedSCAM improved final accuracy from 56.57\% to 59.38\% (+2.81\%), and for ResNet-18 at $\alpha=0.5$ (50 rounds, $E=1$), from 58.00\% to 62.60\% (+4.60\%) under the pasted logs. These improvements support the core hypothesis that uniform $\rho$ is suboptimal once client landscapes vary in gradient scale/consensus.
\begin{table}[b]
\centering
\caption{Final test accuracy on Fashion-MNIST ($\alpha=0.1$) after 10 communication rounds (ResNet-18).}
\label{tab:fmnist-final-acc}
\begin{tabular}{lc}
\toprule
\textbf{Algorithm} & \textbf{Final Accuracy (\%)} \\
\midrule
FedSCAM & 65.41 \\
FedLESAM & 65.76 \\
FedNoLoWe & 57.24 \\
FedWMSAM & 52.68 \\
FedLW & 46.53 \\
FedLWSAM & 42.01 \\
\bottomrule
\end{tabular}
\end{table}

\subsection{Comparison with Aggregation Baselines}

\begin{figure}
\centering
\includegraphics[width=\linewidth]{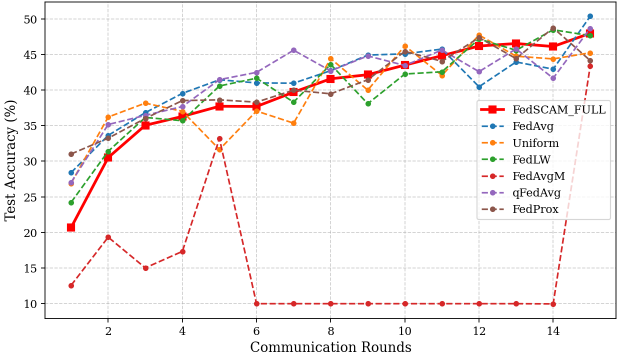}
  \vspace{-20pt}
\caption{Aggregation-focused baselines on CIFAR-10 ($\alpha=0.1$).}
\label{fig:cifar10-agg-alpha01}
\end{figure}

To contextualize the contribution of FedSCAM’s \emph{aggregation} component independent of local optimizer choices, we perform ablations using standard SGD local training. We compare \textbf{FedSCAM (WA)} (which applies our heterogeneity and alignment-based weighting to standard SGD updates, without the personalized SAM component) against widely used aggregation strategies including FedAvg, FedAvgM, q-FedAvg, and FedLW. 

The results in \Cref{tab:cifar10-final-acc} highlight a critical insight: in regimes of moderate heterogeneity ($\alpha=0.5$), the aggregation strategy is the dominant factor in global convergence. FedSCAM (WA) achieves a test accuracy of \textbf{70.63\%}, surpassing both the canonical FedAvg (\textbf{69.68\%}) and the recent loss-weighting baseline FedLW (\textbf{69.42\%}). This suggests that downweighting clients with high gradient variance (high heterogeneity scores) effectively filters out "noisy" updates that would otherwise destabilize the global model.

Furthermore, \cref{fig:cifar10-agg-alpha01} also reveals some stability advantages over momentum-based aggregation. As shown in the training dynamics for $\alpha=0.1$ (and also observed for $\alpha=0.5$), \textbf{FedAvgM} suffers from catastrophic divergence, collapsing to random guessing ($\sim$10\% accuracy) for several consecutive rounds (e.g., rounds 6–13) before recovering. This indicates that server-side momentum can amplify conflicting update directions under label skew. In contrast, FedSCAM (WA) maintains robust, monotonic convergence throughout training. Even under extreme heterogeneity ($\alpha=0.1$), FedSCAM (WA) reaches \textbf{48.39\%}, offering a substantial improvement over robust baselines like FedProx (\textbf{44.14\%}) and demonstrating that intelligent client reweighting can mitigate client drift as effectively as proximal regularization without altering the local objective.

\subsection{Combined Analysis}

\begin{figure}[t]
\centering
\includegraphics[width=\linewidth]{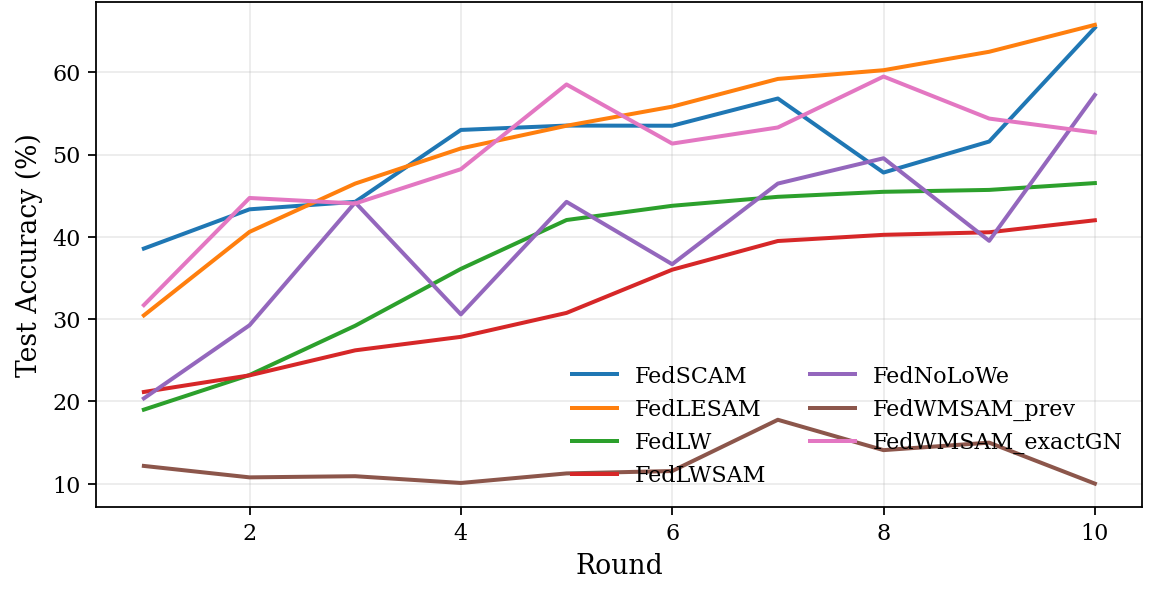}
  \vspace{-20pt}
\caption{Accuracy curves on Fashion-MNIST ($\alpha=0.1$).}
\label{fig:fmnist-alpha01-acc}
\end{figure}

\begin{figure}[t]
\centering
\includegraphics[width=\linewidth]{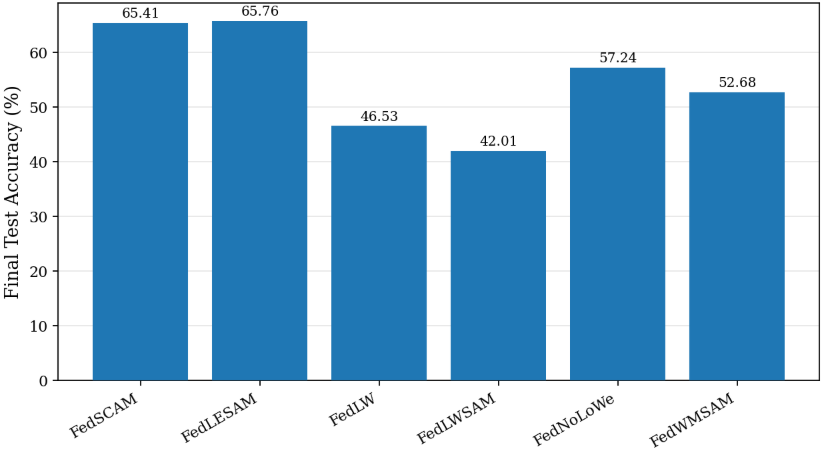}
  \vspace{-20pt}
\caption{Final accuracy distribution on Fashion-MNIST ($\alpha=0.1$). FedSCAM achieves competitive accuracy with FedLESAM, while taking less time.}
\label{fig:fmnist-alpha01-acc-final}
\end{figure}

In the highly heterogeneous distributions, neither advanced aggregation nor sharpness-aware minimization alone is sufficient to guarantee optimal performance. Our combined analysis on Fashion-MNIST ($\alpha=0.1$) shows that FedSCAM successfully bridges this gap, occupying a distinct "sweet spot" in the optimization landscape.

The comparative results in \Cref{tab:fmnist-final-acc} demonstrate that the combination of adaptive radii and heterogeneity-aware aggregation yields performance superior to the sum of its parts. Pure aggregation methods like FedLW struggle to model the complex local landscapes, achieving only \textbf{46.53\%}, while non-adaptive SAM methods like FedWMSAM suffer from instability, reaching only \textbf{52.68\%}. By simultaneously throttling perturbations on noisy clients and downweighting their aggregation influence, FedSCAM achieves \textbf{65.41\%}, effectively recovering the performance of the computationally more expensive FedLESAM (\textbf{65.76\%}).

As illustrated in \Cref{fig:fmnist-alpha01-acc}, FedSCAM avoids the catastrophic convergence failures observed in baselines such as FedAvgM. In extreme non-IID settings, conflicting gradients often cause momentum-based methods to oscillate or even collapse to random guessing. FedSCAM’s alignment-aware update ($h_i^{\mathrm{adj}}$) acts as a dynamic stabilizer: it allows the model to learn from heterogeneous data when the directional signal is consistent ($c_i > 0$) but dampens the update when the signal conflicts with the global trajectory. This results in the tight accuracy distribution shown in \Cref{fig:fmnist-alpha01-acc-final}, indicating that FedSCAM is not only accurate on average but reliable across different initialization seeds and data partitions.

While FedLESAM marginally outperforms FedSCAM by $0.35\%$ in final accuracy on Fashion-MNIST, this comes at a significant computational cost. Our timing benchmarks (\Cref{fig:fmnist-fedscam-time}) indicate that FedSCAM achieves $99.4\%$ of FedLESAM's accuracy while being $\sim12\%$ faster, likely due to the experiment being run serially on the Colab server. This establishes FedSCAM as a highly practical alternative for resource-constrained federated environments.

\begin{table}
\centering
\caption{Final Test Accuracy on CIFAR-10 across varying heterogeneity levels ($\alpha$) after 15 Communication Rounds.}
\label{tab:cifar10-final-acc}
\begin{tabular}{lccc}
\toprule
\textbf{Algorithm} & \textbf{$\alpha=0.1$} & \textbf{$\alpha=0.5$} & \textbf{$\alpha=1.0$} \\
\midrule
FedAvg & 50.36 & 69.68 & 71.16 \\
Uniform & 45.23 & 69.02 & 69.62 \\
FedProx & 44.14 & 69.48 & 71.59 \\
q-FedAvg & 48.67 & 69.06 & 70.16 \\
FedAvgM & 43.39 & 58.03 & 58.90 \\
FedLW & 47.64 & 69.42 & 70.64 \\
FedSAM & 46.92 & 65.09 & 65.40 \\
FedLESAM & 46.25 & 60.62 & 67.68 \\
FedWMSAM & 37.39 & 50.63 & 50.28 \\
FedLWSAM & 44.53 & 64.68 & 64.79 \\
\midrule
\textbf{FedSCAM (WA)} & 48.39 & \textbf{70.63} & 70.63 \\
\textbf{FedSCAM (SAM)} & \textbf{48.77} & 64.65 & 66.50 \\
\textbf{FedSCAM (Full)} & 47.97 & 65.41 & 65.51 \\
\bottomrule
\end{tabular}
\end{table}

\begin{figure}[t]
\centering
\includegraphics[width=\linewidth]{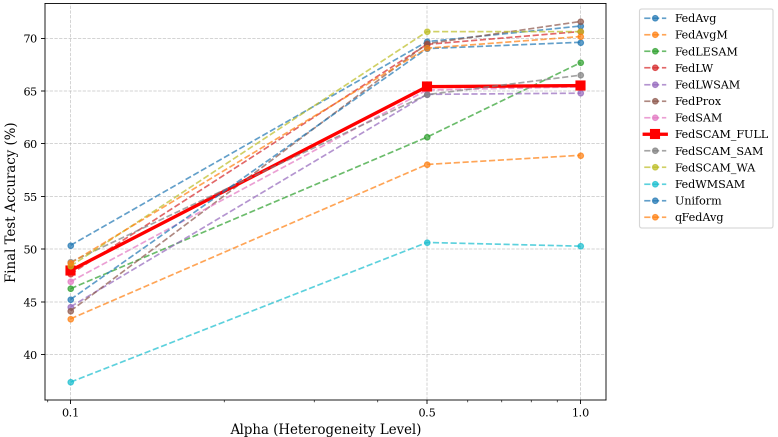}
  \vspace{-20pt}
\caption{Performance of all methods vs.\ $\alpha$ on CIFAR-10 (final accuracy summary). FedSCAM performs better comparatively in more heterogenous settings.}
\label{fig:cifar10-all-alpha}
\end{figure}

\subsection{Client Drift Analysis}

We quantify client drift by measuring the average L2 deviation of local models from the pre-aggregation global model at each round:
\[
d_t = \frac{1}{|\mathcal{S}_t|}\sum_{i\in\mathcal{S}_t} \|w_i - w_t\|_2.
\]

\begin{figure}
\centering
\includegraphics[width=0.85\linewidth]{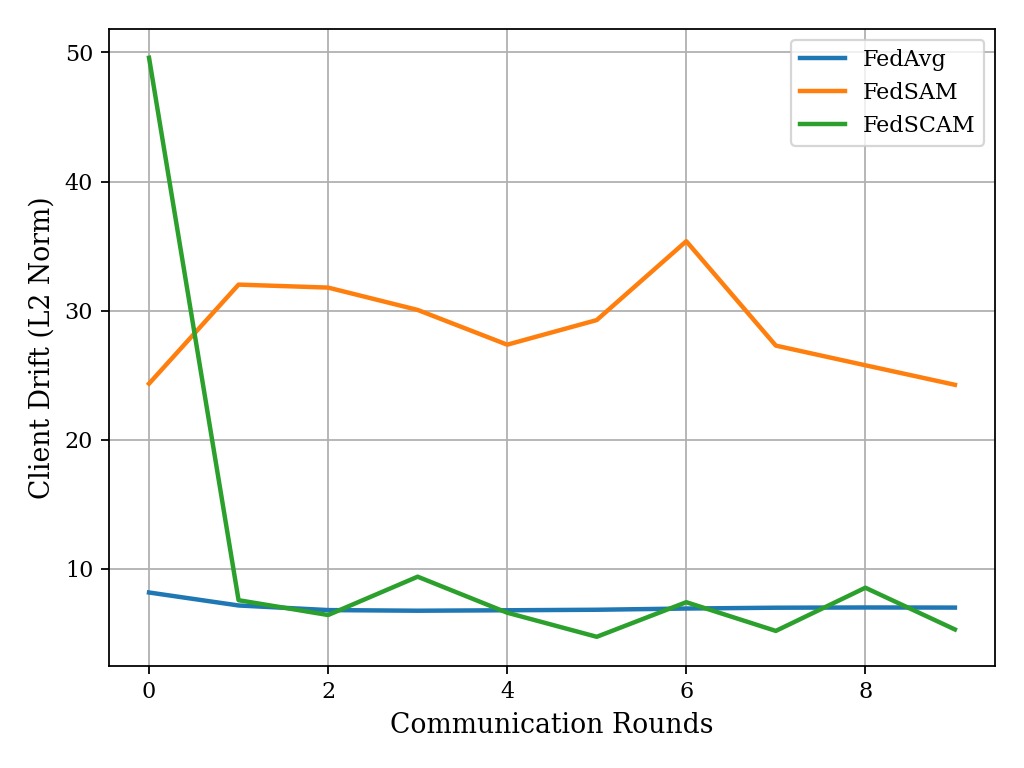}
  \vspace{-17pt}
\caption{Mean client drift per round on CIFAR-10 with $\alpha=0.1$ for FedAvg, FedSAM, and FedSCAM}
\label{fig:clientdrift}
\end{figure}

Figure \ref{fig:clientdrift} illustrates the client drift dynamics over the training horizon. We observe that \textbf{FedSAM} exhibits the highest drift, consistently diverging further from the global model than the other methods. This empirical evidence confirms that blindly applying a fixed perturbation radius in heterogeneous environments exacerbates local divergence, effectively pushing clients toward disjoint local minima. 

In contrast, \textbf{FedSCAM} maintains a low drift profile comparable to FedAvg. This stability validates our hypothesis: by modulating the perturbation radius $\rho_i$ based on geometric alignment and down-weighting conflicting updates, FedSCAM successfully leverages sharpness-aware minimization without suffering from the "drift-amplification" characteristic of standard FedSAM.

\section{Discussion}
\label{sec:discussion}

\begin{figure}[t]
\centering
\includegraphics[width=\linewidth]{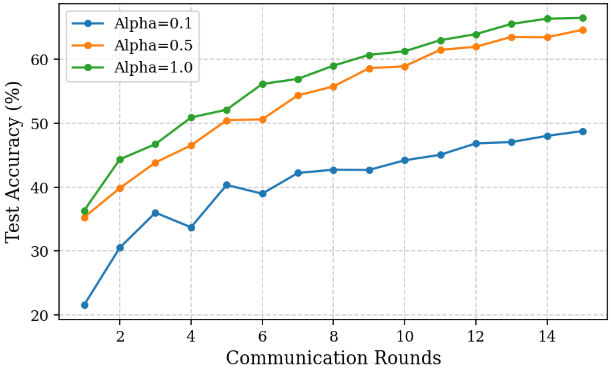}
  \vspace{-20pt}
\caption{Ablation: FedSCAM-SAM accuracy vs.\ rounds on CIFAR-10.}
\label{fig:cifar10-fedscamsam}
\end{figure}

\begin{figure}
\centering
\includegraphics[width=\linewidth]{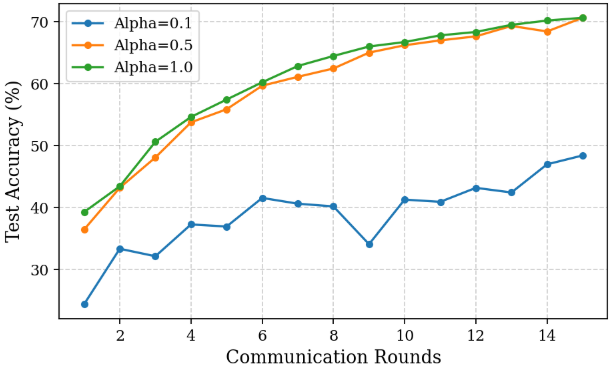}
  \vspace{-20pt}
\caption{Ablation: FedSCAM-WA accuracy vs.\ rounds on CIFAR-10.}
\label{fig:cifar10-fedscamwa}
\end{figure}

The experimental results validate our central thesis: \emph{in the presence of statistical heterogeneity, the benefits of sharpness-aware minimization can only be fully realized if the perturbation radius and aggregation strategy are modulated per-client.} Below, we dissect the mechanisms driving FedSCAM's performance.

\subsubsection{Adaptive Radius as a "Trust Throttle."}
Standard FedSAM applies a uniform perturbation radius $\rho$ across the federation. Our results indicate this is suboptimal: a $\rho$ large enough to find flat minima on "clean" clients causes catastrophic gradient explosions on clients with high label skew (noisy loss landscapes). By modulating $\rho_i \propto (1 + \alpha_\rho h_i)^{-1}$, FedSCAM effectively acts as a "trust throttle." It permits aggressive exploration (large $\rho$) on clients with stable gradients, enforcing flatness where it is safe to do so. Conversely, for clients exhibiting high variance (large $h_i$), it constrains the optimization to a tighter trust region (small $\rho$), preventing the local model from drifting into sharp, non-generalizable valleys. The mean-radius plot (\Cref{fig:fmnist-fedscam-rho}) confirms this dynamic behavior: the average effective radius adapts over time, relaxing as the global model stabilizes.

\subsubsection{Aggregation: The Dominant Lever in Moderate Skew}
A striking finding from our CIFAR-10 ablation study (\Cref{tab:cifar10-final-acc}) is the strength of \textbf{FedSCAM (WA)}. At $\alpha=0.5$, the aggregation-only variant (using standard SGD locally) outperformed FedSAM and FedProx, achieving \textbf{70.63\%}. This suggests that in regimes of moderate heterogeneity, \emph{filtering} conflicting updates is often more impactful than \emph{fixing} local optimization trajectories. Unlike FedLW, which weights based on training loss—a metric that conflates "hard samples" with "noisy samples"—FedSCAM weights based on gradient variance and directional alignment. This distinction is crucial: a client may have high loss because it holds difficult, informative examples (desirable), but if it has high gradient variance and negative alignment, it is likely effectively an outlier (undesirable). FedSCAM correctly downweights the latter, acting as a soft, inexpensive filter against client drift.

\subsubsection{The Efficiency-Robustness Pareto Frontier}
While FedLESAM remains a state-of-the-art baseline, our analysis highlights a critical trade-off. FedLESAM relies on estimating a global perturbation direction, which typically incurs extra communication or memory overheads to synchronize gradients. In our implementation, this manifested as a $\sim$12\% increase in wall-clock time per round compared to FedSCAM. Although FedLESAM achieved marginally higher final accuracy on Fashion-MNIST (+0.35\%), FedSCAM offers a more attractive Pareto point for resource-constrained edge environments. It achieves comparable flatness-seeking benefits using only local signals ($h_i$) and lightweight server-side alignment checks ($c_i$), avoiding the "coordination tax" of global perturbation estimation.

\subsubsection{Stability vs. Momentum}
The collapse of some baselines like FedAvgM in high-skew settings ($\alpha=0.1$) serves as a cautionary tale for aggregation-based FL. Server-side momentum accumulates historical gradients; when client distributions are highly non-IID, this history becomes stale and contradictory to the current local gradients, amplifying oscillations. FedSCAM avoids this trap by using an instantaneous alignment check ($c_i = \cos(s_i, u_{t-1})$). Because this modulation happens "live" at aggregation time, it prevents the accumulation of contradictory signals. The tight final accuracy distribution in \Cref{fig:fmnist-alpha01-acc-final} serves as empirical evidence of this enhanced stability.

\subsubsection{Limitations and Future Work}
FedSCAM introduces computational overhead primarily through the calculation of heterogeneity scores ($h_i$) and pilot directions. While these costs are amortized over local epochs, they may be non-negligible for extremely lightweight IoT devices. Future work could explore: (1) cheaper proxies for $h_i$, such as tracking weight changes rather than gradient norms, and (2) theoretical bounds connecting the alignment-aware aggregation weights to the convergence rate of non-convex federated optimization. Additionally, while the clustered dampening step helps in extreme skew, its optimal configuration ($K, \lambda$) currently requires manual tuning; adaptive clustering remains an open direction.

\section{Conclusion}
\label{sec:conclusion}

We have presented FedSCAM, a robust optimization framework that fundamentally rethinks how Sharpness-Aware Minimization is applied in heterogeneous federated environments. By dynamically modulating the perturbation radius based on local gradient variance and enforcing alignment-aware aggregation, FedSCAM effectively neutralizes the destabilizing effects of non-IID data without incurring the computational overhead of global gradient estimation. Our extensive empirical evaluation on Fashion-MNIST and CIFAR-10 confirms that FedSCAM achieves a competitive performance in accuracy, stability, and efficiency compared to existing state-of-the-art baselines. Ultimately, this work demonstrates that heterogeneity should not be treated merely as noise to be averaged out, but as a critical signal for modulating trust, establishing FedSCAM as a scalable foundation for resilient federated learning in practical scenarios.


\appendix
\clearpage
\section*{Appendix}

\textbf{Additional Implementation Details}\\
For the Dirichlet distribution generation, we utilized the standard implementation provided in most FL benchmarks. We ensured that each client received a minimum number of samples (min\_size=10) to prevent numerical instabilities during training. The heterogeneity score calculation was performed on the first 3 batches of the local data loader to keep computational overhead low.

\begin{figure}[H]
\centering
\includegraphics[width=0.85\linewidth]{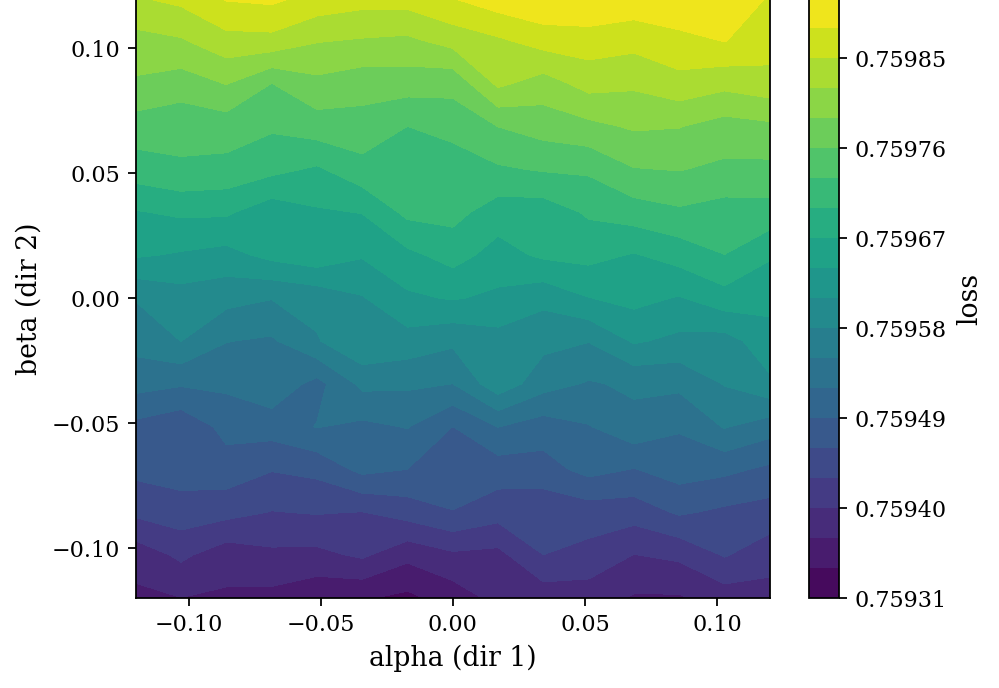}
  \vspace{-16pt}
\caption{2D Loss Landscape visualization for FedSAM on Fashion-MNIST. FedSAM has a smooth, featureless landscape, as all the clients calculate perturbaton locally, so the global model becomes the average of several disjoint minima.}
\label{fig:fmnist-fedsam-2d}
\end{figure}

\begin{figure}[H]
\centering
\includegraphics[width=0.85\linewidth]{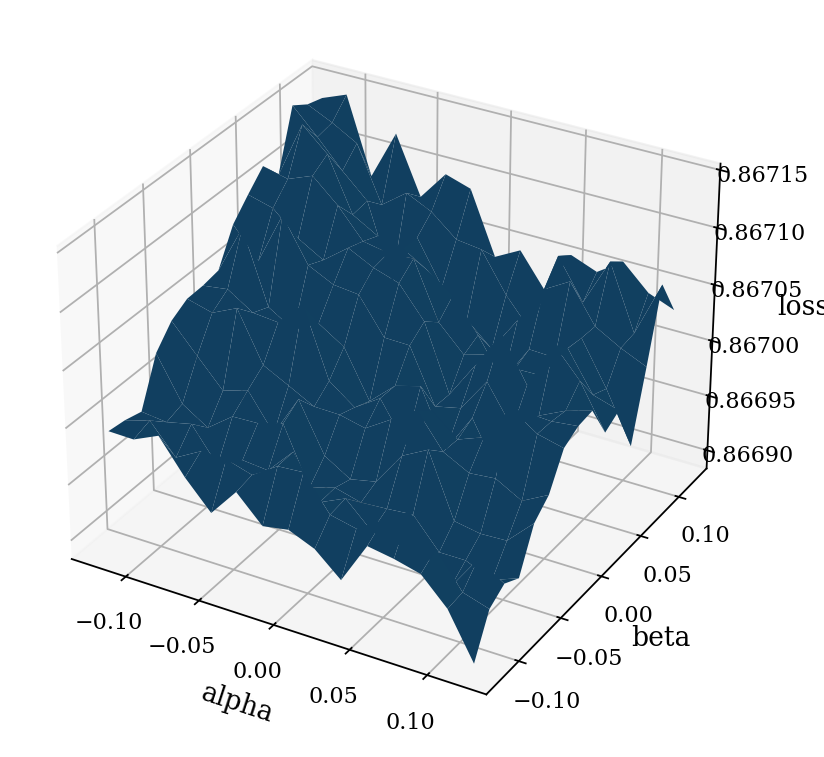}
  \vspace{-16pt}
\caption{3D Loss Landscape visualization for FedSCAM on Fashion-MNIST, clearly showing jagged minima and peaks.}
\label{fig:fmnist-fedscam-3d}
\end{figure}

\begin{figure}[H]
\centering
\includegraphics[width=0.8\linewidth]{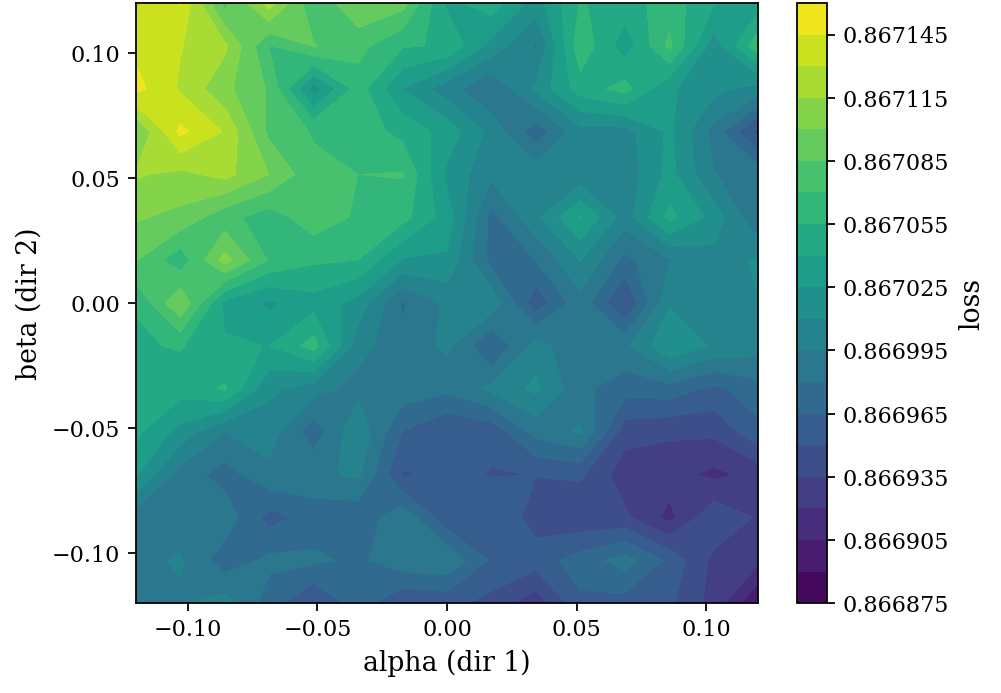}
  \vspace{-15pt}
\caption{2D Loss Landscape visualization for FedSCAM on Fashion-MNIST. FedSCAM manages to capture more of the global minima and spikes compared to FedSAM.}
\label{fig:fmnist-fedscam-2d}
\end{figure}

\begin{figure}[H]
\centering
\includegraphics[width=0.8\linewidth]{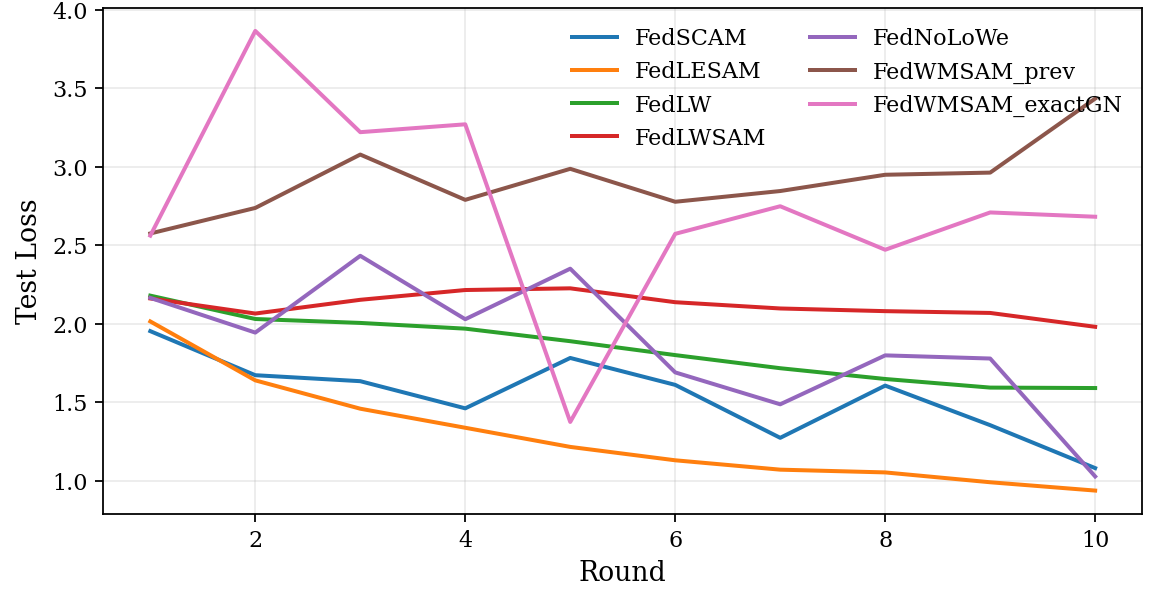}
  \vspace{-15pt}
\caption{Loss metrics on Fashion-MNIST ($\alpha=0.1$). FedSCAM is competitive with FedLESAM and beats the other algorithms.}
\label{fig:fmnist-alpha01-loss}
\end{figure}

\begin{figure}[H]
\centering
\includegraphics[width=\linewidth]{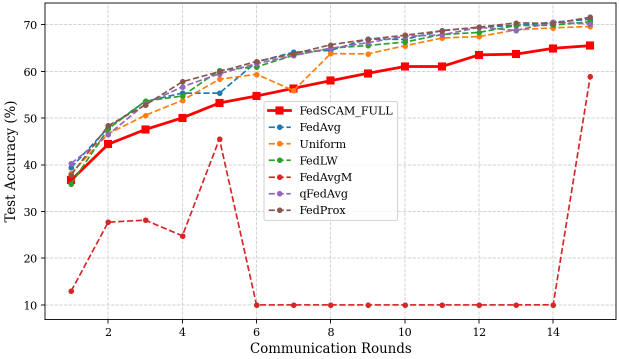}
  \vspace{-15pt}
\caption{Aggregation variants on CIFAR-10 ($\alpha=1.0$).}
\label{fig:cifar10-agg-alpha1}
\end{figure}

\begin{figure}[H]
\centering
\includegraphics[width=\linewidth]{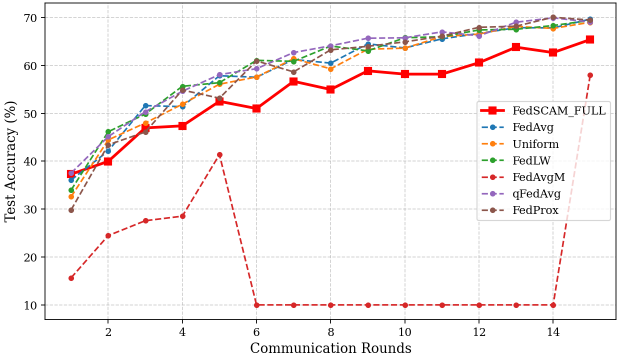}
  \vspace{-15pt}
\caption{Aggregation variants on CIFAR-10 ($\alpha=0.5$).}
\label{fig:cifar10-agg-alpha05}
\end{figure}

\begin{figure}
\centering
\includegraphics[width=\linewidth]{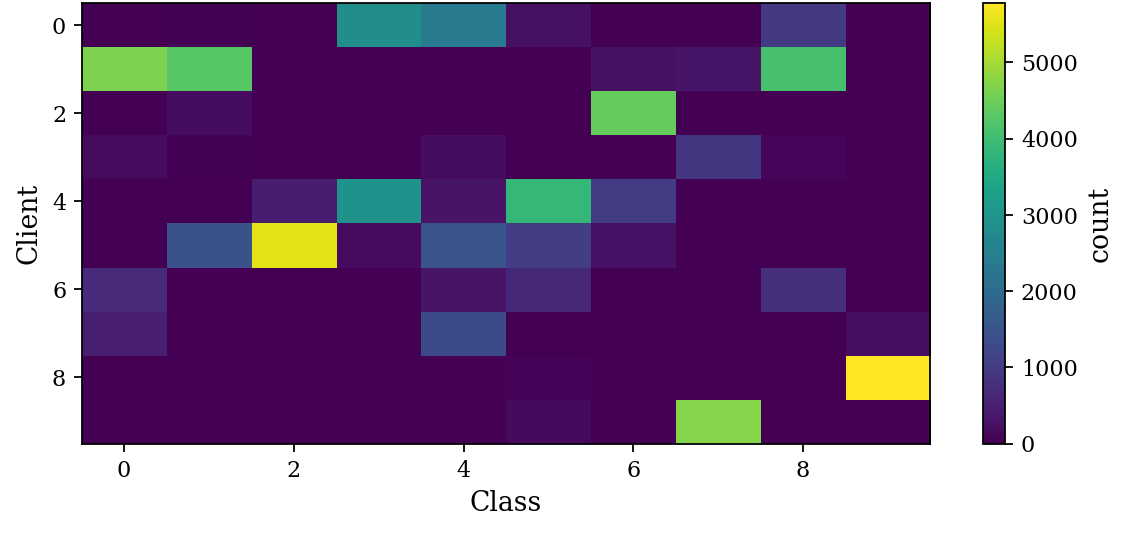}
  \vspace{-15pt}
\caption{Client label distributions for Fashion-MNIST (illustrative non-IID partition).}
\label{fig:fmnist-data-dist}
\end{figure}

\begin{figure}
\centering
\includegraphics[width=0.8\linewidth]{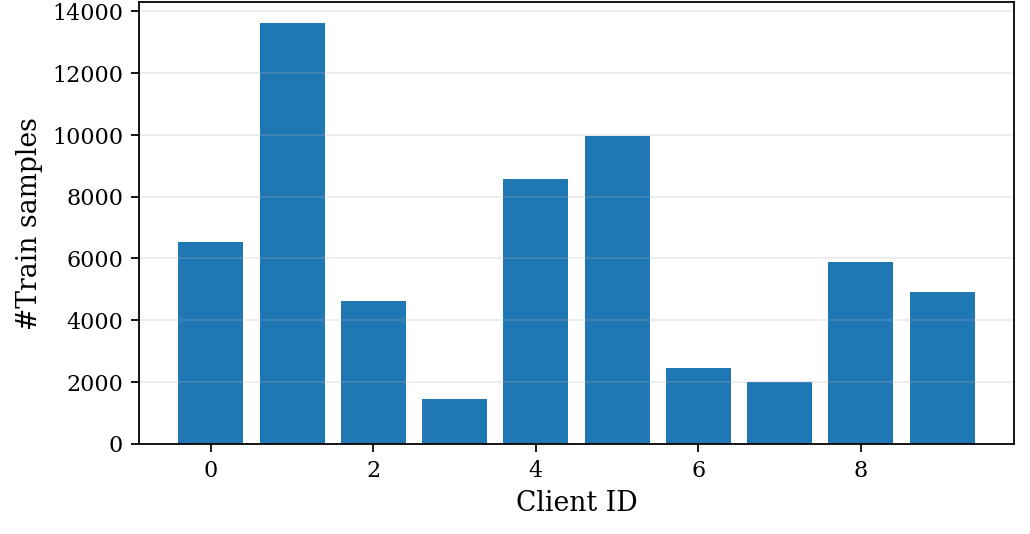}
  \vspace{-15pt}
\caption{Client dataset sizes for Fashion-MNIST. Under extreme heterogeneity conditions, some clients carry much more samples to train from.}
\label{fig:fmnist-data-size}
\end{figure}

\begin{figure}[H]
\centering
\includegraphics[width=0.8\linewidth]{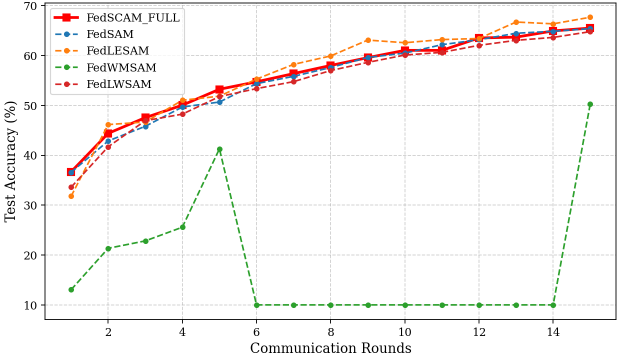}
  \vspace{-15pt}
\caption{SAM variants on CIFAR-10 ($\alpha=1.0$).}
\label{fig:cifar10-sam-alpha1}
\end{figure}

\begin{figure}[H]
\centering
\includegraphics[width=0.8\linewidth]{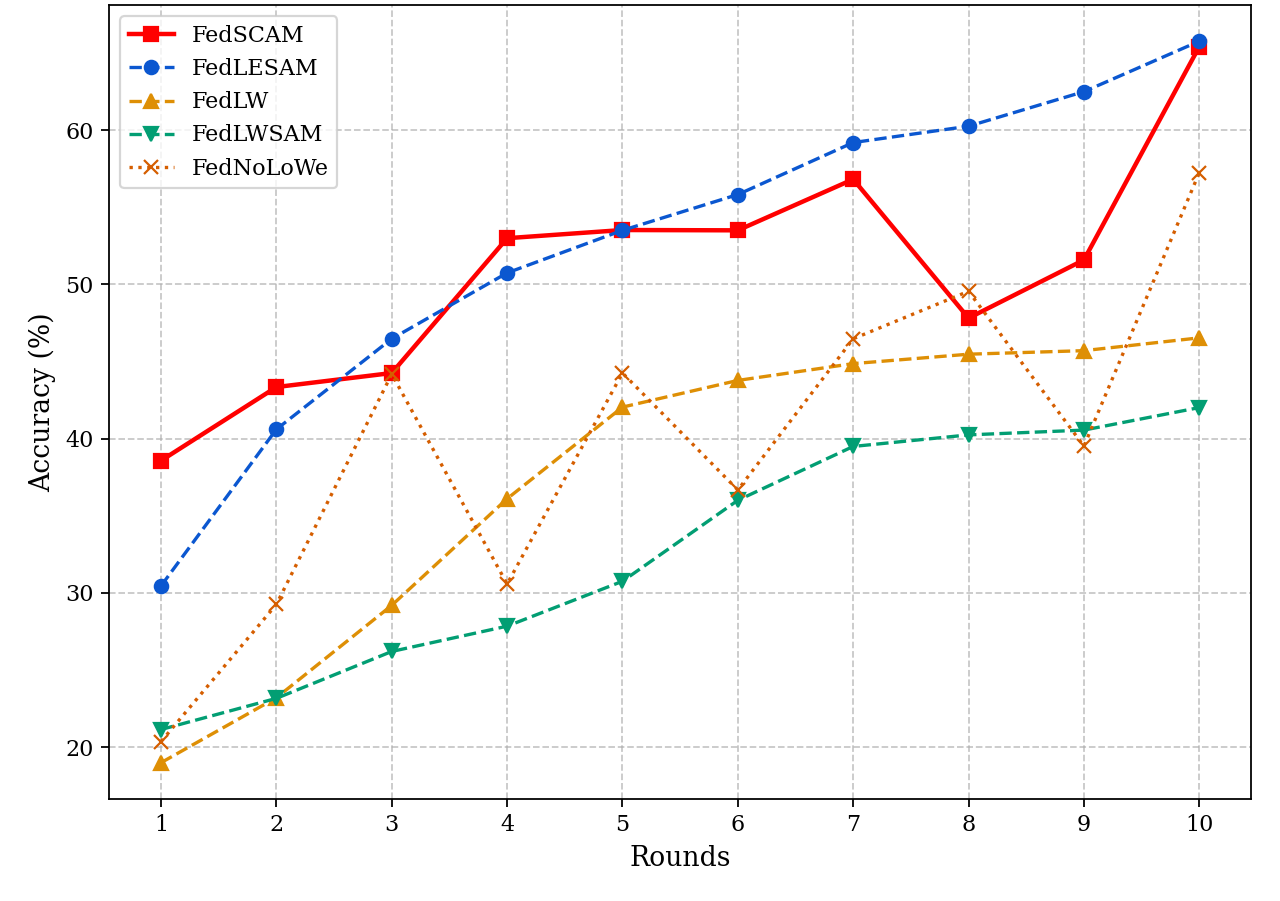}
  \vspace{-15pt}
\caption{Comparison: FedSCAM vs. FedSAM on Fashion-MNIST.}
\label{fig:fmnist-fedscam-vs-sam}
\end{figure}

\begin{figure}[H]
\centering
\includegraphics[width=0.8\linewidth]{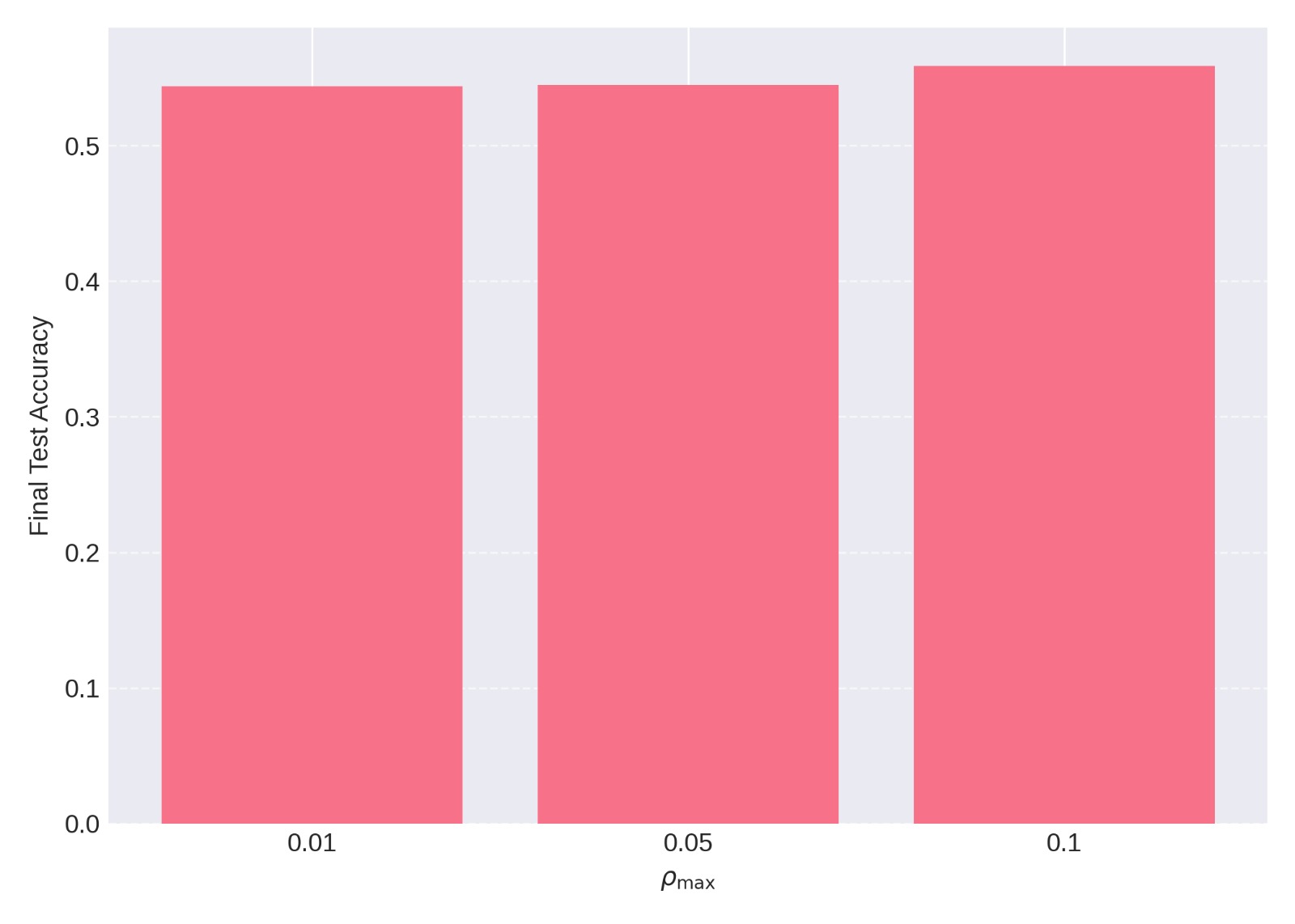}
  \vspace{-15pt}
\caption{Impact of $\rho_{\max}$ on Final Test Accuracy (Fashion-MNIST), showing a positive correlation.}
\label{fig:rhomax-final}
\end{figure}

\begin{figure}[H]
\centering
\includegraphics[width=0.8\linewidth]{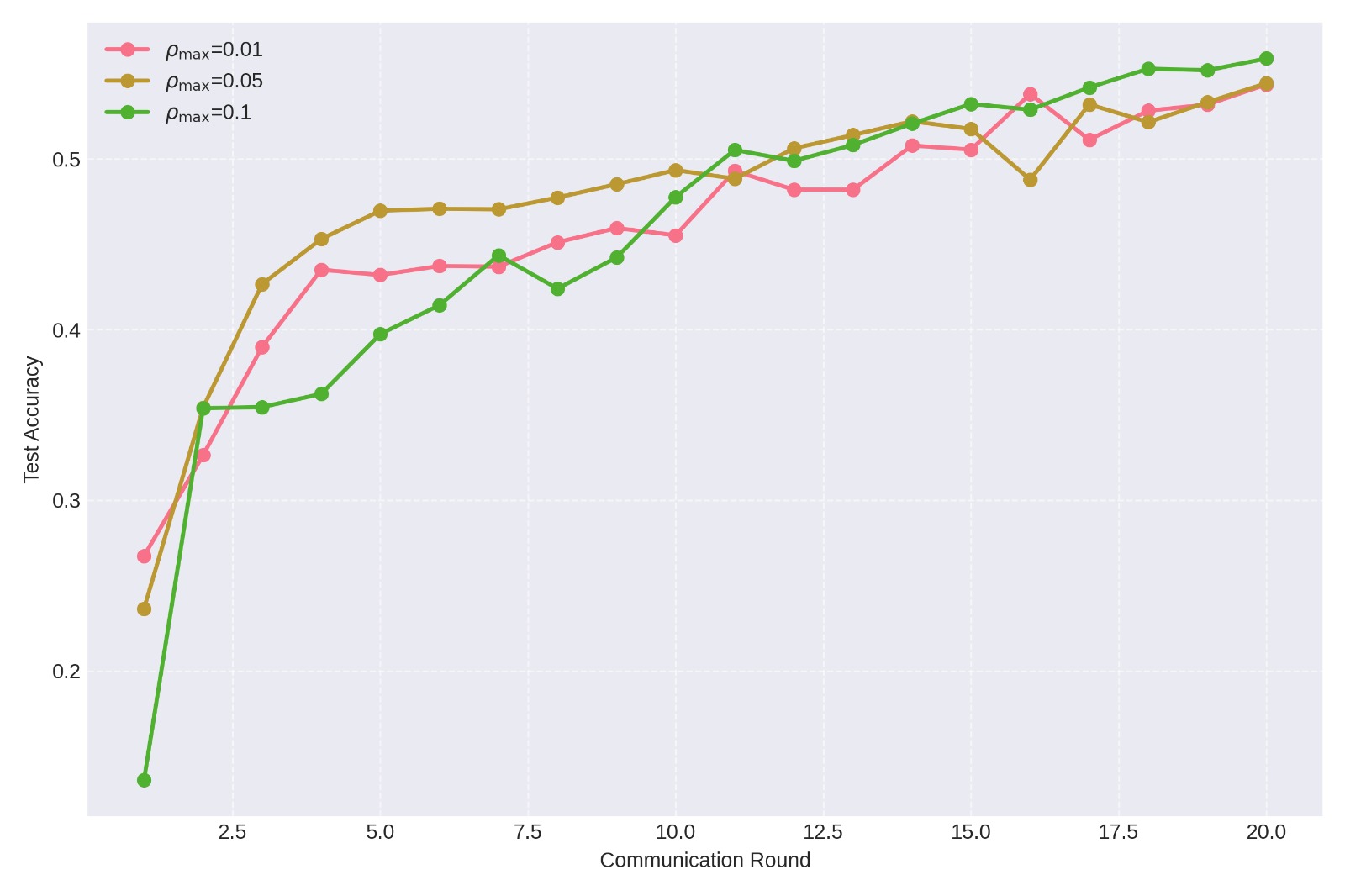}
  \vspace{-15pt}
\caption{Test Accuracy vs. Rounds for different $\rho_{\max}$ values.}
\label{fig:rhomax-rounds}
\end{figure}


\begin{figure}[H]
\centering
\includegraphics[width=0.8\linewidth]{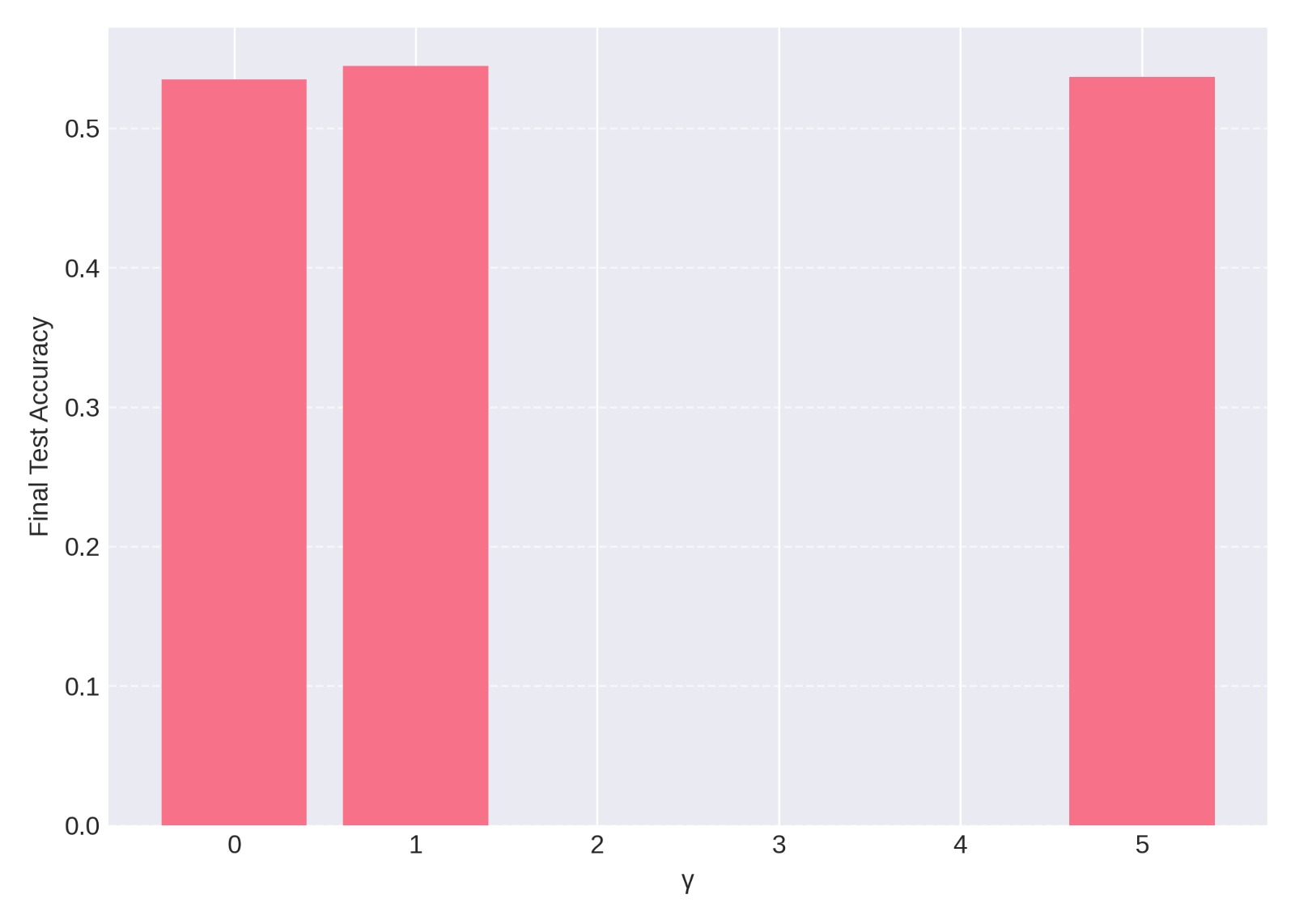}
  \vspace{-15pt}
\caption{Impact of Heterogeneity Penalty $\gamma$ on Final Test Accuracy.}
\label{fig:gamma-final}
\end{figure}

\begin{figure}[H]
\centering
\includegraphics[width=0.8\linewidth]{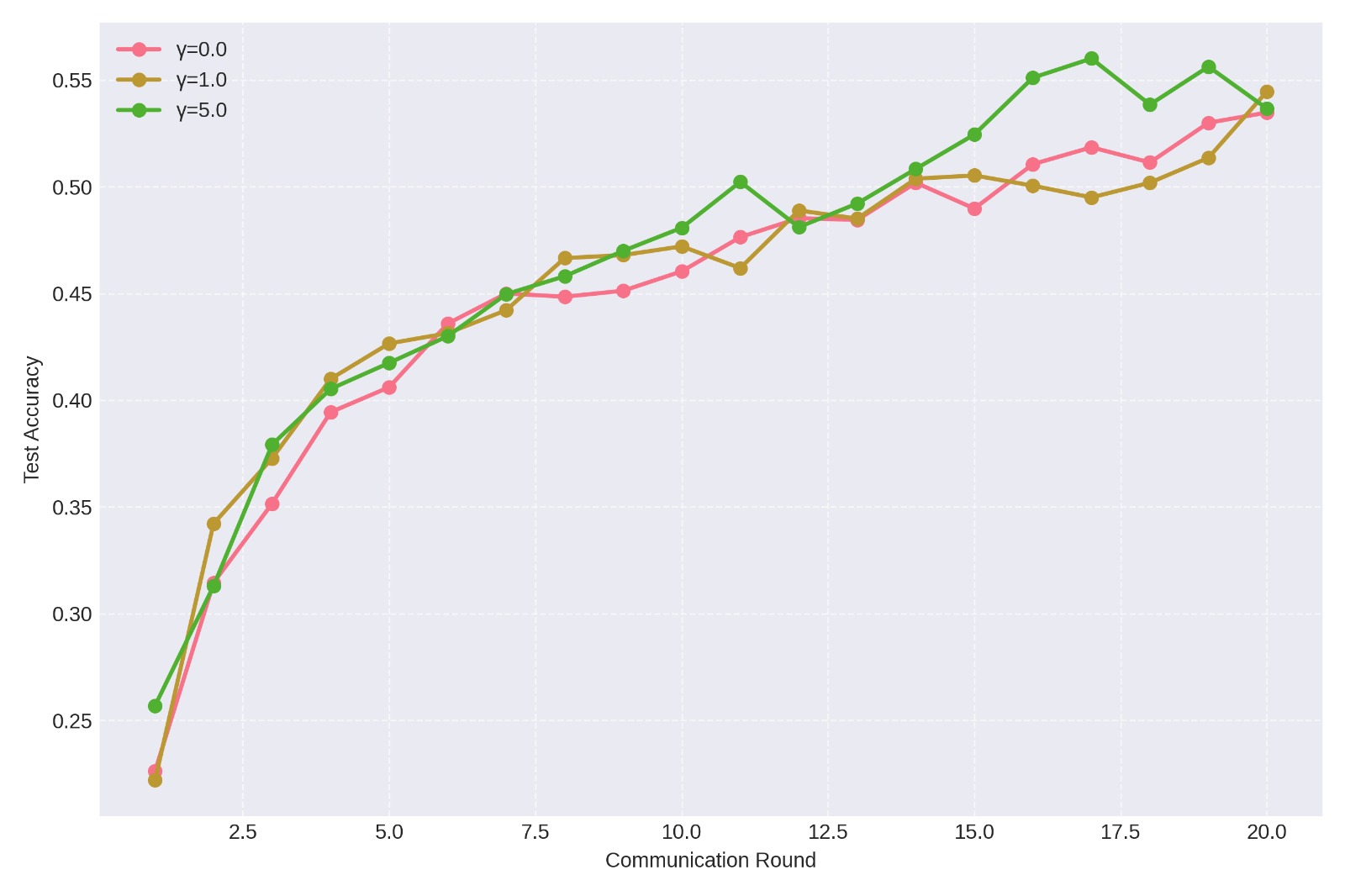}
  \vspace{-15pt}
\caption{Test Accuracy vs. Rounds for different $\gamma$ values.}
\label{fig:gamma-rounds}
\end{figure}


\begin{figure}[H]
\centering
\includegraphics[width=0.8\linewidth]{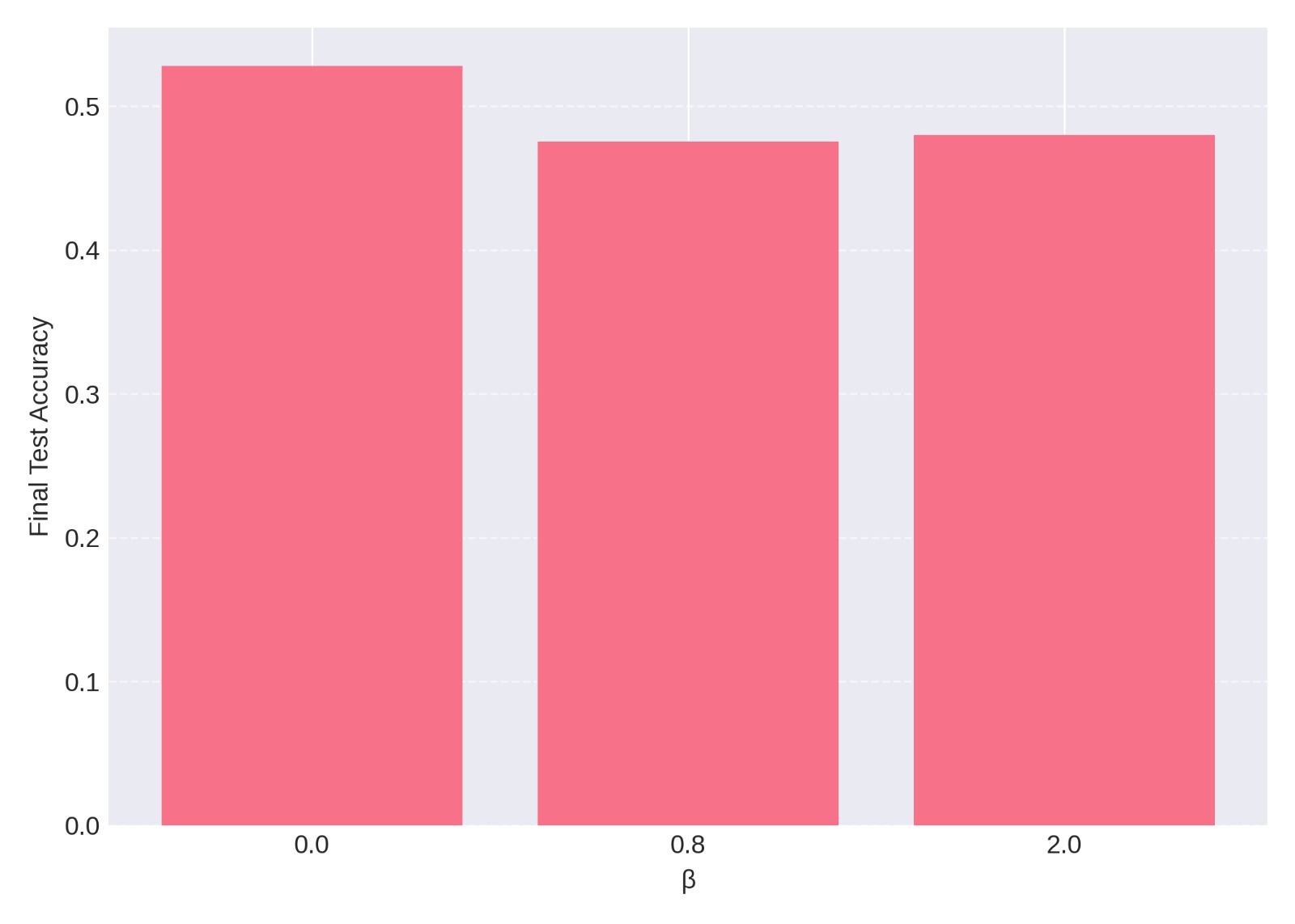}
  \vspace{-15pt}
\caption{Impact of Alignment Boost $\beta$ on Final Test Accuracy.}
\label{fig:beta-final}
\end{figure}

\begin{figure}[H]
\centering
\includegraphics[width=0.8\linewidth]{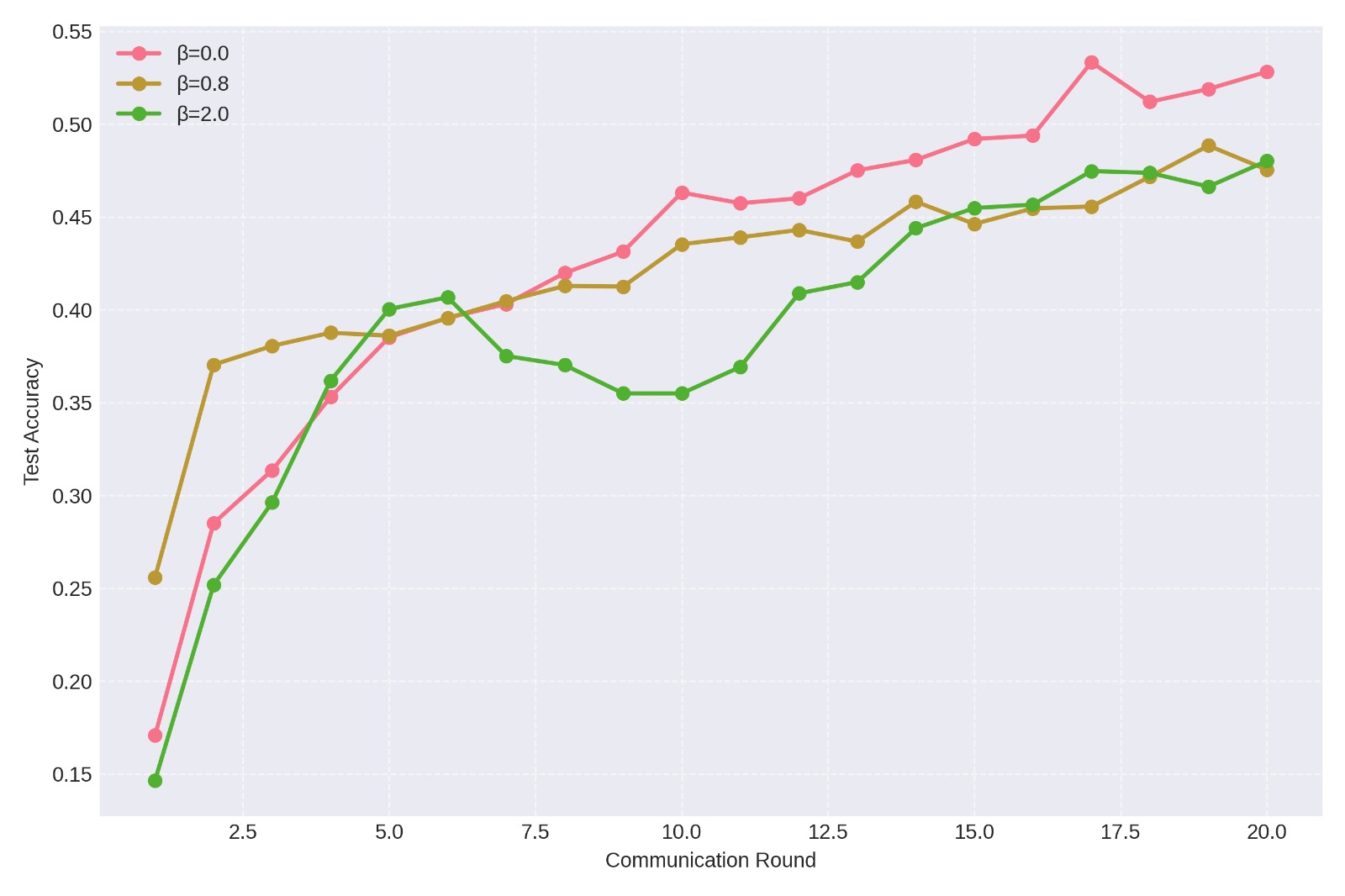}
  \vspace{-15pt}
\caption{Test Accuracy vs. Rounds for different $\beta$ values.}
\label{fig:beta-rounds}
\end{figure}


\small
\begin{thebibliography}{99}

\bibitem[McMahan et~al.(2017)]{mcmahan2017fedavg}
H.~B. McMahan, E.~Moore, D.~Ramage, S.~Hampson, and B.~A. y Arcas.
\newblock Communication-efficient learning of deep networks from decentralized data.
\newblock In \emph{Proceedings of the 20th International Conference on Artificial Intelligence and Statistics (AISTATS)}, 2017.

\bibitem[Foret et~al.(2021)]{foret2021sharpness}
P.~Foret, A.~Kleiner, H.~Mobahi, and B.~Neyshabur.
\newblock Sharpness-aware minimization for efficiently improving generalization.
\newblock In \emph{International Conference on Learning Representations (ICLR)}, 2021.
\newblock arXiv:2010.01412.

\bibitem[Qu et~al.(2022)]{qu2022fedsam}
Z.~Qu, X.~Li, R.~Duan, Y.~Liu, B.~Tang, and Z.~Lu.
\newblock Generalized federated learning via sharpness aware minimization.
\newblock In \emph{Proceedings of the 39th International Conference on Machine Learning (ICML)}, 2022.
\newblock arXiv:2206.02618.

\bibitem[Fan et~al.(2024)]{jiang2024fedlesam}
Z.~Fan, S.~Hu, J.~Yao, G.~Niu, Y.~Zhang, M.~Sugiyama, and Y.~Wang.
\newblock Locally estimated global perturbations are better than local perturbations for federated sharpness-aware minimization.
\newblock \emph{arXiv preprint arXiv:2405.18890}, 2024.

\bibitem[Li et~al.(2025)]{li2025fedwmsam}
T.~Li, Y.~Huang, L.~Jiang, C.~Liu, Q.~Xie, W.~Du, L.~Wang, and K.~Wu.
\newblock A fast and flat federated learning method via weighted momentum and sharpness-aware minimization.
\newblock \emph{arXiv preprint arXiv:2511.22080}, 2025.

\bibitem[Hsu et~al.(2019)]{hsu2019measuring}
T.-M.~H. Hsu, H.~Qi, and M.~Brown.
\newblock Measuring the effects of non-identical data distribution for federated visual classification.
\newblock \emph{arXiv preprint arXiv:1909.06335}, 2019.

\bibitem[Li et~al.(2020a)]{li2020qfedavg}
T.~Li, M.~Sanjabi, A.~Beirami, and V.~Smith.
\newblock Fair resource allocation in federated learning.
\newblock In \emph{International Conference on Learning Representations (ICLR)}, 2020.
\newblock arXiv:1905.10497.

\bibitem[Li et~al.(2020b)]{li2020fedprox}
T.~Li, A.~K. Sahu, M.~Zaheer, M.~Sanjabi, A.~Talwalkar, and V.~Smith.
\newblock Federated optimization in heterogeneous networks.
\newblock In \emph{Proceedings of Machine Learning and Systems (MLSys)}, 2020.
\newblock arXiv:1812.06127.

\bibitem[Yao and Wang(2025)]{yao2025fedlw}
C.~Yao and W.~Wang.
\newblock A loss-based weighted aggregation method for federated learning with heterogeneous computing resources.
\newblock In \emph{Proceedings of the 2025 International Conference on Computing and Artificial Intelligence}, 2025.
\newblock DOI: 10.1145/3735358.3737773.

\bibitem[Le et~al.(2025)]{le2025fednolowe}
D.-D. Le, T.-N. Huynh, A.-K. Tran, M.-S. Dao, and P.~T. Bao.
\newblock FedNolowe: A normalized loss-based weighted aggregation strategy for robust federated learning in heterogeneous environments.
\newblock \emph{PLOS ONE}, 20(8):e0322766, 2025.

\end{thebibliography}
\end{document}